\documentclass{article}

\usepackage[preprint]{neurips_2022}

\usepackage[utf8]{inputenc} 
\usepackage[T1]{fontenc}    
\usepackage{hyperref}       
\usepackage{booktabs}       
\usepackage{amsfonts}       
\usepackage{nicefrac}       
\usepackage{microtype}      
\usepackage{xcolor}         
\usepackage{graphicx}
\usepackage{multirow}
\usepackage{natbib}
\usepackage{amsmath}
\usepackage{amsthm}
\usepackage{algorithm}
\usepackage{algpseudocode}
\usepackage{amssymb}
\newtheorem{theorem}{Theorem}
\newtheorem{corollary}{Corollary}
\newtheorem{lemma}{Lemma}

\newtheorem{assumption}{Assumption}

\DeclareMathOperator*{\argmin}{\arg\!\min}
\newcommand{\iid}{ {\stackrel{i.i.d.}{\sim}} }
\makeatletter
\newcommand{\setalgname}[1]{%
  \renewcommand{\ALG@name}{#1} 
}
\makeatother
\allowdisplaybreaks

\title{Model uncertainty quantification using feature confidence sets for outcome excursions}

\author{%
  Junting Ren \\
  Division of Biostatistics\\
  University of California San Diego\\
  \texttt{j5ren@ucsd.edu} \\
  \And
  Armin Schwartzman \\
  Division of Biostatistics and Hal{\i}c{\i}o\u{g}lu Data Science Institute,\\ 
  University of California San Diego \\
  \texttt{armins@ucsd.edu} \\
}

\begin{document}

\maketitle

\begin{abstract}
  When implementing prediction models for high-stakes real-world applications such as medicine, finance, and autonomous systems, quantifying prediction uncertainty is critical for effective risk management. Traditional approaches to uncertainty quantification, such as confidence and prediction intervals, provide probability coverage guarantees for the expected outcomes $f(\boldsymbol{x})$ or the realized outcomes $f(\boldsymbol{x})+\epsilon$. Instead, this paper introduces a novel, model-agnostic framework for quantifying uncertainty in continuous and binary outcomes using confidence sets for outcome excursions, where the goal is to identify a subset of the feature space where the expected or realized outcome exceeds a specific value. The proposed method constructs data-dependent inner and outer confidence sets that aim to contain the true feature subset for which the expected or realized outcomes of these features exceed a specified threshold. We establish theoretical guarantees for the probability that these confidence sets contain the true feature subset, both asymptotically and for finite sample sizes. The framework is validated through simulations and applied to real-world datasets, demonstrating its utility in contexts such as housing price prediction and time to sepsis diagnosis in healthcare. This approach provides a unified method for uncertainty quantification that is broadly applicable across various continuous and binary prediction models.
\end{abstract}

\section{Introduction}
The responsible development and deployment of prediction models in high-stakes real-life applications, such as medicine \citep{begoli2019need}, stock portfolio management \citep{christoffersen2011elements}, and autonomous driving \citep{michelmore2020uncertainty}, require quantification of prediction uncertainty for effective risk management. Uncertainty quantification of prediction models involves different methods depending on the nature of the outcomes being predicted.

For continuous outcomes, model prediction uncertainty is typically quantified by constructing confidence or prediction bands $L$ and $U$, such that
\begin{equation*}
    \mathbb{P}\left[L(\hat{f}(\boldsymbol{x})) \leq f(\boldsymbol{x}) \leq U(\hat{f}(\boldsymbol{x}))\right]\geq 1-\alpha
\end{equation*}
or
\begin{equation*}
\mathbb{P}\left[L(\hat{f}(\boldsymbol{x})) \leq f(\boldsymbol{x})+\epsilon \leq U(\hat{f}(\boldsymbol{x}))\right]\geq 1-\alpha.
\end{equation*}

Here, $\alpha$ is the pre-defined type-I error, $f(\boldsymbol{x})$ represents a deterministic function for the data point with feature vector $\boldsymbol{x}$, and $L$ and $U$ are the lower and upper limits of a confidence or prediction band, often functions of the point prediction $\hat{f}(\boldsymbol{x})$ for the \textit{realized outcome} $f(\boldsymbol{x})+\epsilon$. The term $\epsilon$ refers to irreducible noise with an expectation of zero. Since $\mathbb{E}\left[\hat{f}(\boldsymbol{x})+\epsilon\right]=f(\boldsymbol{x})$ if the point prediction is unbiased, we call $f(\boldsymbol{x})$ the \textit{expected outcome}. For binary outcomes, $f(\boldsymbol{x})$ is directly the probability of occurrence of the outcome.

For example, in linear regression, the point prediction is $\hat{f}(\boldsymbol{x})=\boldsymbol{x}\boldsymbol{\hat\beta}$, and the lower and upper limits of the confidence intervals can be determined by adding and subtracting from the point estimate $\hat{f}(\boldsymbol{x})$ the estimated standard deviation at $\boldsymbol{x}$, multiplied by a certain percentile of the Normal or T-distribution \citep{rencher2008linear}, by bootstrapping, or a combination of both. For neural networks, the variance and confidence intervals are computed using deterministic methods \citep{van2020uncertainty,lee2020gradients}, Bayesian approaches \citep{blundell2015weight, rezende2015variational}, or ensemble techniques \citep{lakshminarayanan2017simple}. Simultaneous confidence intervals have been developed to maintain coverage rates for more than one observation in generalized linear regression \citep{sun1994simultaneous, liu2008construction}.


\subsection{Uncertainty quantification for excursion sets}
Confidence and prediction bands guarantee the probability of containing the expected or realized outcome. Instead, we are interested in uncertainty quantification in the feature space. Specifically, we wish to identify feature data points in the test set for which we can confidently assert that \textit{all} their corresponding outcomes \textit{simultaneously} exceed a specific value. For instance, in emergency rooms, doctors may need to know which subset of patients with certain features will have a high probability of developing sepsis within six hours to apply early treatment, and also which patients are unlikely to develop sepsis within the same timeframe to allocate healthcare resources efficiently \citep{seymour2017time}. In the context of stock and portfolio management, when constructing a diversified portfolio, managers might want to select a subset of assets that exceed a specific return threshold under various market conditions to achieve desired diversification benefits and minimize risk \cite{meucci2009managing}. 

There is a growing consensus on the need to pinpoint the region of features where predicted classifications are uncertain \citep{hamidzadeh2019identification, penrod2022success}, which can occur when a test instance's feature vector values lie close to the decision boundary. Methods for capturing this uncertainty region have been proposed for support vector machines using geometric margin \citep{voichita2008identifying} and for neural networks using adversarial attack \citep{alarab2022adversarial}. However, these methods are not applicable to scenarios with continuous outcomes and lack formal statistical theory. 

This article proposes a novel model-agnostic \textit{inverse inference} framework for quantifying prediction uncertainty. Instead of \textit{forward inference} that captures the uncertainty in the point predictions, we define \textit{inverse inference} as uncertainty quantification in the space of the features (\textit{inverse set} or \textit{feature set}), as we describe below. 

Given a new set of \textit{testing features} without the outcome labels, denoted as $\mathcal{X}_m:=\{ \boldsymbol{x}_i \in \mathbb{R}^d, i = 1,\dots, m\}$, the goal of this approach is to estimate a subset of $\mathcal{X}_m$ defined by the excursion of the expected or realized outcome above $c$:
\begin{equation*}
    \mathcal{X}_m(c) := \{\boldsymbol{x}_i \in \mathcal{X}_m: f(\boldsymbol{x}_i) \geq c\}
\end{equation*}
or
\begin{equation*}
    \mathcal{X}_m(c) := \{\boldsymbol{x}_i \in \mathcal{X}_m: f(\boldsymbol{x}_i)+\epsilon_i \geq c\},
\end{equation*}
where the testing features are treated as fixed. We call $\mathcal{X}_m(c)$ {\em inverse set} or {\em feature set} because it is the preimage or inverse image of a set $[c, +\infty) \subset \mathbb{R}$ under the deterministic function $f: \mathcal{X} \mapsto \mathbb{R}$. The point estimate of such a feature set is $\hat{\mathcal{X}}_m(c) := \{\boldsymbol{x}_i \in \mathcal{X}_m: \hat f_n(\boldsymbol{x}_i) \geq c\}
$. We aim to quantify the uncertainty in this feature set by constructing data-dependent inner and outer feature confidence sets $\mathrm{CS}^{i}_{c}$ and $\mathrm{CS}^{o}_{c}$, which are sub- and super-sets of the feature set with probability at least $1-\alpha$:
\begin{align}
\mathbb{P}\left[
    \mathrm{CS}_c^i \subseteq 
    \mathcal{X}_m(c)
    \subseteq 
    \mathrm{CS}_c^o
\right]\geq 1-\alpha.
    \label{CS_containment}
\end{align}

The $\mathrm{CS}^{i}_{c}$ is the set of feature points where we are confident that their expected or realized outcomes are greater than $c$. The region outside $\mathrm{CS}^{o}_{c}$ is where we are confident that the expected or realized outcomes are less than $c$, and the region outside $\mathrm{CS}^{i}_{c}$ but inside $\mathrm{CS}^{o}_{c}$ is where the model is uncertain whether the predictions are greater or smaller than $c$. 

\begin{figure}
    \centering
    \includegraphics[scale = 0.32]{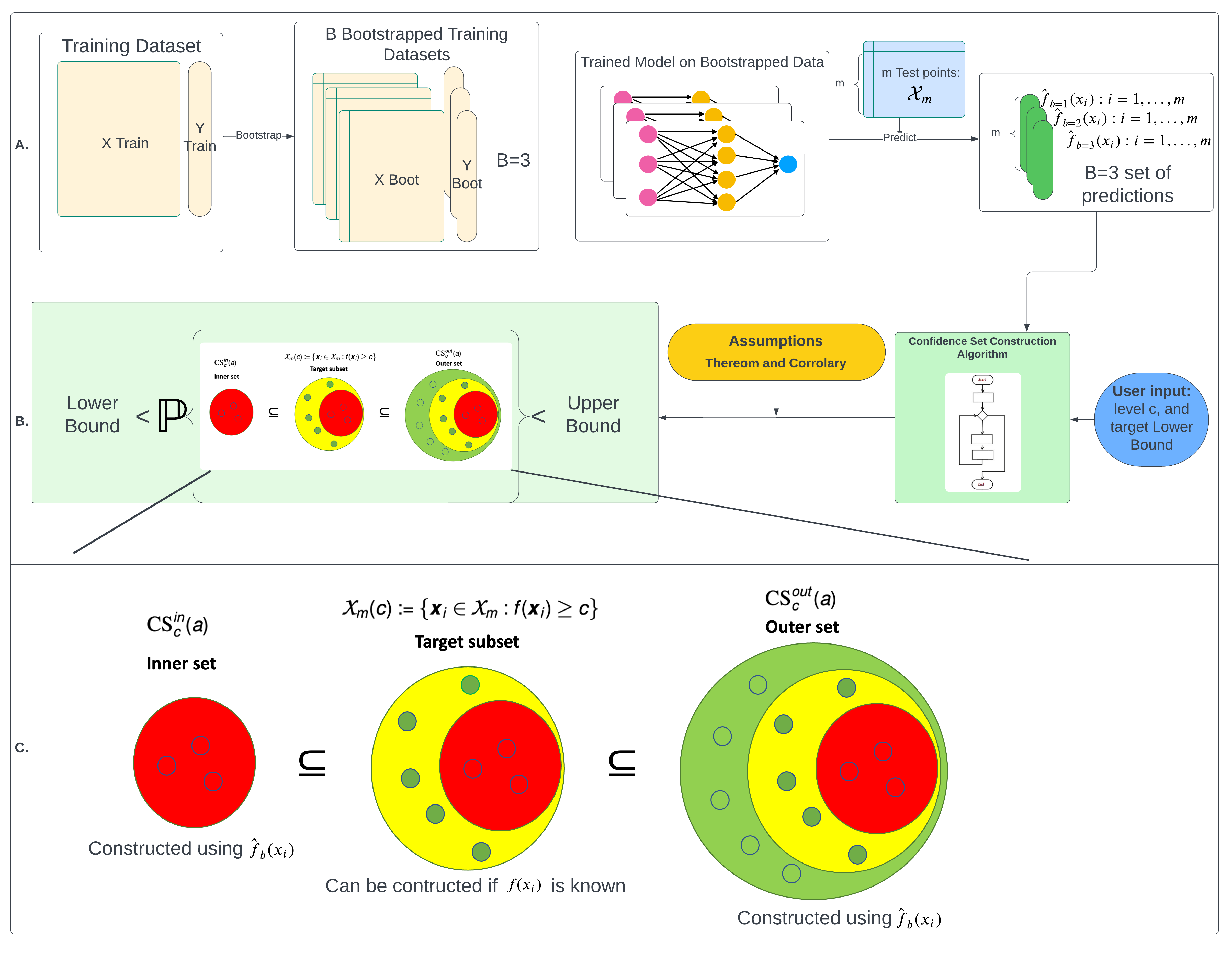}
    \caption{Confidence set construction workflow. A. Given a training data set $(X,Y)$ and a pre-defined model structure, $B$ non-parametric bootstrap training datasets are used to train $B$ models with different weights. Each of the trained models predicts the outcomes for the unseen $m$ test feature points $\mathcal{X}_m$, obtaining $B$ set of predictions. The user specifies the excursion level $c$ and target coverage probability lower bound $1-\alpha$. Based on the $B$ sets of predictions on the $m$ test feature points, the confidence set construction algorithm outputs the estimated inner set $\mathrm{CS}^{i}_{c}$ and the outer set $\mathrm{CS}^{o}_{c}$ with probability guarantees (above the target lower-bound and below the upper-bound estimated by the algorithm) that the true targeted subset is sandwiched within the two estimated sets. The estimated outer set $\mathrm{CS}^{o}_{c}$ (green, yellow and red region) contains the unknown target subset $\mathcal{X}_m(c)$ (red and yellow region). Subsequently, the target subset contains the estimated inner set $\mathrm{CS}^{i}_{c}$ (red region). We are confident that the red points belong to $\mathcal{X}_m(c)$, while the green points remain uncertain. If the assumptions hold, our theorem and corollary guarantee this containment statement is lower and upper bounded.}
    \label{fig:flow_chart}
\end{figure}

The construction of the inner set and the outer set is model-agnostic, as demonstrated in Figure \ref{fig:flow_chart}A. Given any type of model structure, a non-parametric bootstrap is conducted on the training dataset. Based on $m$ unseen new test feature data points, different sets of predictions of the $m$ feature points are generated from the models trained on different bootstrapped training datasets. The confidence set construction algorithm takes in a user-specified level of interest $c$, the targeted containment probability  $1-\alpha$ and the predictions, and outputs the inner and outer set for the $m$ test feature points, as shown in Figure \ref{fig:flow_chart}B. The probability of the containment statement $\mathrm{CS}_c^i \subseteq \mathcal{X}_m(c)\subseteq \mathrm{CS}_c^o$ is lower bounded by $1-\alpha$ and upper bounded by a value estimated by the algorithm, if certain assumptions are satisfied. This relationship is visualized in Figure \ref{fig:flow_chart}C. In other words, if we classify $\mathrm{CS}^{i}_{c}$ as the set where the expected or realized outcomes are greater than $c$ (positive set), we achieve 100\% precision since all selected points are true positives, with probability at least $1-\alpha$. If we classify $\mathrm{CS}^{o}_{c}$ as the positive set, we obtain 100\% sensitivity since the selected points contain all the true positive points, with probability at least $1-\alpha$. The samples in $\mathrm{CS}^{o}_{c}$ but outside of $\mathrm{CS}^{i}_{c}$ are the samples with uncertain classification. 

\begin{figure}
    \centering
    \includegraphics[scale=0.6]{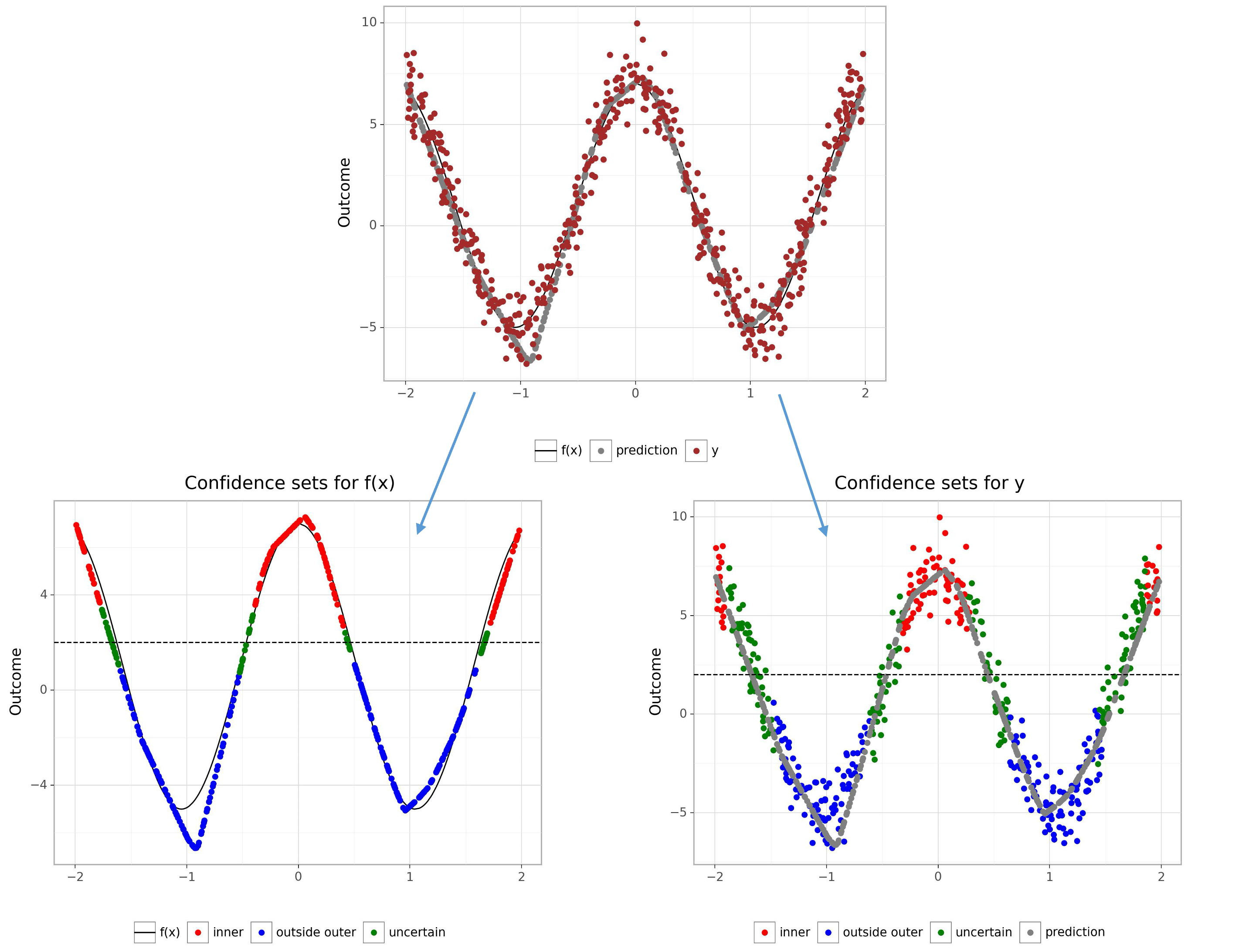}
    \caption{Demonstration of confidence sets constructed for 500 testing points in a 1D space, using a two-layer neural network model trained on 100 training samples where the true function is a cosine function. Upper plot: the expected outcome $f(x)$ (solid black line), predictions for 500 points (gray), and the realized outcomes $y$ (red). Left plot: confidence sets for the expected outcome $f(x)$ at the level $c=2$. The inner set $\mathrm{CS}^{i}_{c}$ (red) and  the outer set $\mathrm{CS}^{o}_{c}$ (red + green) are labeled on the point predictions; the complement of the outer set is labeled in blue. Right plot: confidence sets for the realized outcome $y$ with similar labeling.}
    \label{fig:demo_CS}
\end{figure}

The example in Figure \ref{fig:demo_CS} demonstrates how confidence sets can help identify test feature points whose expected or realized outcomes are greater than or less than a threshold $c$ (red or blue points, respectively) and where the points are uncertain (green points) with respect to the specific model structure and data instance. This enables decision-makers to have a better understanding of the uncertainty associated with the model's predictions, facilitating more informed decisions in real-world applications.

\subsection{Contributions}
This paper makes the following contributions:
\begin{itemize}
    \item A novel model-agnostic framework for quantifying uncertainty for continuous or binary outcomes through confidence sets in feature space for outcome excursions.
    \item A theoretical guarantee that the probability for the confidence set containment statement in Equation \eqref{CS_containment} for \textit{expected outcomes} converges to pre-defined probability for asymptotic training sample sizes or is sandwiched between the target lower and estimated upper bound for finite training samples. 
    \item Confidence sets construction algorithms that ensure that the probability of the containment statement in Equation \eqref{CS_containment} is lower and upper bounded for both \textit{expected} and \textit{realized outcomes}. The validity of the algorithms are demonstrated in simulations where the model structure is correctly or incorrectly specified.
    \item The algorithm and simulation code are available on \url{https://github.com/junting-ren/model_uncertainty_using_feature_confidence_set}.
\end{itemize}

\subsection{Existing confidence set construction methods}
In previous work \citep{ren2024inverse}, it was shown that the confidence sets $\mathrm{CS}^{i}_{c}$ and $\mathrm{CS}^{o}_{c}$ can be obtained by simply inverting lower and upper simultaneous confidence intervals (SCIs). However, that work focused on constructing confidence sets for \textit{model parameters} rather than prediction. Also, that approach requires access to methods to construct valid SCIs, which not all prediction models have. Thus, in this paper, we focus on constructing confidence sets for prediction uncertainty quantification and formulating theories that guarantee non-conservative coverage rates non-asymptotically. By fixing the level $c$, rather than requiring simultaneous control over all levels, the proposed approach yields more power in detecting significant points compared to directly inverting SCIs, resulting in a larger inner set $\mathrm{CS}^{i}_{c}$ and a smaller outer set $\mathrm{CS}^{o}_{c}$.

Other existing inverse set estimation methods focus on low-dimensional continuous dense domains such as 1D density data \citep{saavedra2016comparative}, 3D brain images \citep{bowring2019spatial}, and 2D maps \citep{sommerfeld2018confidence}. These methods often require strong assumptions such as the underlying deterministic function to be continuous \citep{sommerfeld2018confidence,qiao2019nonparametric}, the mean function to be Gaussian \citep{bolin2015excursion,french2013spatio} and the coverage probability is only asymptotic for large sample size \citep{sommerfeld2018confidence, qiao2019nonparametric}. However, in building models for prediction, the dimension $p$ of the feature vector $\boldsymbol{x}$ can be up to hundreds of thousands with a relatively small number of data points inside the high dimensional domain $\mathbb{R}^p$ \citep{clarke2008properties,sert2020analysis}. This high dimensional non-dense setup essentially invalidates the approaches stated above.

\subsection{Outline}
Section \ref{sec:theory} presents the main theoretical results for both asymptotically and non-asymptotically single-level $c$ confidence sets containment probability statements. Section \ref{sec:algorithm} provides an efficient algorithm to construct the confidence sets with probability guarantee. Section \ref{sec:experiment} demonstrates the validity of our theory and algorithm in simulations for scenarios where the model structure is correctly or incorrectly specified. Section \ref{sec:application} showcases the application of the framework in two real data analyses for housing price and time to sepsis prediction. Section \ref{sec:discussion} concludes with a brief discussion.

\section{Theory}\label{sec:theory}
In this section, we provide theoretical results for constructing confidence sets for the \textit{expected outcome $f(\boldsymbol{x})$}. Theoretical results are not provided for the \textit{realized outcome}, but construction algorithms for both the expected and realized outcomes are provided and validated through simulation.
\subsection{Setup}
Suppose we are given \textit{training data} $(\tilde{\mathcal{X}}_n, \tilde{\mathcal{Y}}_n) := \{(\tilde{\boldsymbol{x}}_i,\tilde{y}_i), \tilde{\boldsymbol{x}}_i \in  \mathbb{R}^p, \tilde{y}_i \in \mathbb{R}, i = 1,\dots,n \}$ where $\tilde{\boldsymbol{x}}_i$ is a vector of features and $\tilde{y}_i$ is the realized outcome, sampled from a joint distribution. Assume there is an underlying deterministic function $f: \mathbb{R}^p \mapsto \mathbb{R}$ such that $y = f(\boldsymbol{x}) +\epsilon$, where $\epsilon$ is the irreducible error of the data which is a random variable with expectation of zero. After constructing a function $\hat f_n: \mathbb{R}^p \mapsto \mathbb{R} $ approximating $f$ from the training data $(\tilde{\mathcal{X}}_n, \tilde{\mathcal{Y}}_n)$, we are given a new set of \textit{testing feature values} $\mathcal{X}_m:=\{ \boldsymbol{x}_i \in \mathbb{R}^p, i = 1,\dots, m\}$ without the corresponding outcome variable values. Using $\hat f_n$, we can obtain a similar tuple $(\mathcal{X}_m, \hat{\mathcal{Y}}_m)=\{(\boldsymbol{x}_i, \hat f(\boldsymbol{x}_i)), i = 1,\dots, m\}$. The goal is to estimate a subset of $\mathcal{X}_m$
$$\mathcal{X}_m(c) := \{\boldsymbol{x}_i \in \mathcal{X}_m: f(\boldsymbol{x}_i) \geq c\},$$
where the testing features are treated as fixed. The point estimate of $\mathcal{X}_m(c)$ is constructed with $\hat f_n$ trained on random sample $(\tilde{\mathcal{X}}_n, \tilde{\mathcal{Y}}_n)$
\begin{equation}
    \hat{\mathcal{X}}_m(c) := \{\boldsymbol{x}_i \in \mathcal{X}_m: \hat f_n(\boldsymbol{x}_i) \geq c\} \label{eq:estimated_feature_set}
\end{equation}

Notice, we use $\boldsymbol{x}_i$ denote the \textit{testing features} with $i=1,\dots, m$ and $\tilde{\boldsymbol{x}}_i$ denote the \textit{training features} with $i=1,\dots, n$. While this notation convention may seem unconventional, it streamlines our notation by placing emphasis on the testing data, which is the focus of our discussion.

\begin{assumption}\label{assump:1}
    There exist a sequence of numbers $\tau_n \rightarrow 0$ and a continuous function $\sigma:\mathcal{X}_m \rightarrow \mathbb{R}^+$ bounded from above and below, and a positive integer $n_0$ such that $\forall n\geq n_0$
    $$
    G_n(\boldsymbol{x}_i) := \frac{\hat f_n(\boldsymbol{x}_i)-f(\boldsymbol{x}_i)}{\tau_n\sigma(\boldsymbol{x}_i)}
    $$
    is a almost surely bounded random field on the discrete domain $ \mathcal{X}_m$ with mean zero, unit variance. That is 
    $$
    \mathbb{P}\left(\sup_{\boldsymbol{x}_i \in \mathcal{X}_m}|G_n(\boldsymbol{x}_i)| < \infty \middle | 
     \boldsymbol{x}_i\right)=1.
    $$
\end{assumption}
The above probability is conditioned on the testing features. We suppress the conditioning in the notation hereafter for brevity. 

\subsection{Feature confidence sets}
To quantify the uncertainty of the estimated set (\ref{eq:estimated_feature_set}), we introduce the feature inner and outer confidence sets 
\begin{align*}
    &\mathrm{CS}^{i}_{c}:=\mathrm{CS}^{in}_{c}(a):= \left\{\boldsymbol{x}_i \in \mathcal{X}_m: \frac{\hat f_n(\boldsymbol{x}_i)-c}{\tau_n\sigma(\boldsymbol{x_i})}\geq a\right\},\\
    &\mathrm{CS}^{o}_{c}:=\mathrm{CS}^{out}_{c}(a)= \left\{\boldsymbol{x}_i \in \mathcal{X}_m: \frac{\hat f_n(\boldsymbol{x}_i)-c}{\tau_n\sigma(\boldsymbol{x_i})}\geq -a\right\}.
\end{align*}

We also introduce a signed distance for the test feature vector $\boldsymbol{x}_i$ to the boundary level $c$ associated with $f$ as 
\begin{align*}
    d_n(\boldsymbol{x}_i) :=
    d^c_n(\boldsymbol{x}_i) :=  \frac{f(\boldsymbol{x}_i)-c}{\tau_n\sigma(\boldsymbol{x}_i)}.
\end{align*}
We define $\boldsymbol{x}_i$ as a boundary point if $d_n(\boldsymbol{x}_i)=0$. Observe that $\mathrm{CS}^{i}_{c}(a)$ and $\mathrm{CS}^{o}_{c}(a)$ are functions of $a$, whose value can be used to control the probability containment statement as shown in the next section.

\subsection{Asymptotic probability statement when boundary points exist}
Consider the test feature space $\mathcal{X}_m \in \mathbb{R}^d$, with a dense grid of test feature points $\boldsymbol{x_i}$ such that $\exists \boldsymbol{x_i} \in \mathcal{X}_m, d_n(\boldsymbol{x_i})=0$. Under these conditions, the following asymptotic probability statement holds.
\begin{theorem}\label{theorem:asy_main_bound}
    If assumption \ref{assump:1} holds, and the set $\{\boldsymbol{x}_i \in \mathcal{X}_m: d_n(\boldsymbol{x}_i)=0\}$ is non-empty then 
    \begin{align*}
        \quad\lim_{n\rightarrow \infty}
        \mathbb{P}\Bigg( 
         \mathrm{CS}_c^i \subseteq 
            \mathcal{X}_m(c)
            \subseteq 
            \mathrm{CS}_c^o
        \Bigg)= \lim_{n\rightarrow \infty}
        \mathbb{P}\left(
            \inf_{\boldsymbol{x}_i \in \{d_n(\boldsymbol{x}_i)=0\}}G_n(\boldsymbol{x}_i)\geq -a\right).
    \end{align*}    
\end{theorem}
To apply Theorem \ref{theorem:asy_main_bound} in practice, the test dataset $\mathcal{X}_m \in \mathbb{R}^d$ would need to be dense in $\mathbb{R}^d$, or else the probability that $\exists \boldsymbol{x_i} \in \mathcal{X}_m, d_n(\boldsymbol{x_i})=0$ would be essentially zero. As the dimension $p$ of the features increases, it becomes increasingly impossible to satisfy this condition. Furthermore, without knowledge of the true data model $f$, even with dense test data, we can only identify neighbors of the boundary points, provided $\hat{f}$ is well-constructed using finite training sample.

To address this issue, we propose Theorem \ref{theorem:finite_main_bound} shown in the next section.

\subsection{Finite-sample probability statement when boundary points do not exist}
In practice, both the number of training points and test feature points are finite. Therefore, instead of focusing on the elusive boundary test points that may not exist, we utilize the test feature points on an inflated boundary to determine the threshold $a$ for constructing the confidence sets. The inflated boundary is defined as:
\begin{align*}
    &\{0 \leq d_n(\boldsymbol{x}_i)\leq e_1\} \text{ and }\{-e_2 \leq d_n(\boldsymbol{x}_i)< 0\},
\end{align*}
where $e_1>0$ and $e_2>0$. Utilizing this inflated boundary, given a finite training sample $n$, the probability containment statement is bounded by a lower and upper bound, rather than an equality statement.

\begin{theorem}\label{theorem:finite_main_bound}
    Let $e_1$ and $e_2$ be any positive constants such that the sets $\{0 \leq d_n(\boldsymbol{x}_i)\leq e_1\}$, $\{-e_2 \leq d_n(\boldsymbol{x}_i)< 0\}$, $\{d_n(\boldsymbol{x}_i)> e_1\}$ and $\{d_n(\boldsymbol{x}_i) < -e_2\}$ are non-empty. Let CARD denote the cardinality of a set. Then,
    $$
    L
    \leq 
    \mathbb{P}\Bigg( 
        \mathrm{CS}_c^i \subseteq 
        \mathcal{X}_m(c)
        \subseteq 
        \mathrm{CS}_c^o
    \Bigg) 
    \leq
    U,
    $$
    with Upper bound    
    \begin{align}
        & U=\mathbb{P}\left(
        \left\{ 
        \inf_{\boldsymbol{x}_i \in \{0 \leq d_n(\boldsymbol{x}_i)\leq e_1\}}G_n(\boldsymbol{x}_i)\geq -a-\sup_{\boldsymbol{x}_i \in \{0\leq d_n(\boldsymbol{x}_i)\leq e_1\}}d_n(\boldsymbol{x}_i)
        \right\}
        \bigcap \right.\nonumber\\
        &\quad \left.
        \left\{ 
        \sup_{\boldsymbol{x}_i \in \{-e_2 \leq d_n(\boldsymbol{x}_i)< 0\}}G_n(\boldsymbol{x}_i)< a+\sup_{\boldsymbol{x}_i \in \{-e_2 \leq d_n(\boldsymbol{x}_i)< 0\}}|d_n(\boldsymbol{x}_i)|
        \right\}
        \right) \tag{U}\label{eq:U}
    \end{align}
    and Lower bound:
    \begin{align*}
        &L= \mathbb{P}\left(
        \left\{ 
        \sup_{\boldsymbol{x}_i \in \{-e_2 \leq d_n(\boldsymbol{x}_i)< e_1\}}|G_n(\boldsymbol{x}_i)|< a+\inf_{\boldsymbol{x}_i \in \{-e_2 \leq d_n(\boldsymbol{x}_i)\leq e_1\}}|d_n(\boldsymbol{x}_i)|
        \right\}
        \right)+\tag{L.1}\label{eq:L.1}\\
        &\quad \sum_{\forall \boldsymbol{x}_i \in \{d_n(\boldsymbol{x}_i)> e_1\}}\mathbb{P}\Bigg(G_n(\boldsymbol{x}_i)\geq -a-d_n(\boldsymbol{x}_i)\Bigg) +\tag{L.2}\\ 
        & \quad \sum_{\forall \boldsymbol{x}_i \in \{d_n(\boldsymbol{x}_i) < -e_2\}} \mathbb{P}\Bigg(G_n(\boldsymbol{x}_i)< a+|d_n(\boldsymbol{x}_i)|
        \Bigg) - \tag{L.2}\\
        & \quad \mathrm{CARD}\Bigg\{\boldsymbol{x}_i \in \mathcal{X}_m: d_n(\boldsymbol{x}_i)> e_1\Bigg\}- \mathrm{CARD}\Bigg\{\boldsymbol{x}_i \in \mathcal{X}_m: d_n(\boldsymbol{x}_i) < -e_2\Bigg\}.\tag{L.2}\label{eq:L.2}
    \end{align*}
\end{theorem}

Finding the points in the inflated boundary $\{0 \leq d_n(\boldsymbol{x}_i)\leq e_1\}$ and $\{-e_2 \leq d_n(\boldsymbol{x}_i)< 0\}$ offers two advantages: first, the existence of such points is guaranteed as long $e_1$ and $e_2$ are sufficiently large; second, locating the points in the inflated boundary that guarantee the upper and lower probability bounding statement is more practical than finding boundary points with finite samples.

\subsection{Asymptotic probability statement when boundary points do not exist}
If boundary points do not exist, but we can identify which test feature points are closest to the boundary, then an asymptotic probability statement can be made for the containment. For a fixed test data set $\mathcal{X}_m$, we denote the following features that are closest to the boundary as:
\begin{align*}
    &\boldsymbol{x}_{+} = \argmin_{\boldsymbol{x}_i\in\{0 \leq d_n(\boldsymbol{x}_i)\}} d_n(\boldsymbol{x}_i)\\
    &\boldsymbol{x}_{-} = \argmin_{\boldsymbol{x}_i\in\{0 > d_n(\boldsymbol{x}_i)\}} |d_n(\boldsymbol{x}_i)| 
\end{align*}

\begin{corollary}\label{corollary:finite_sharp_bound}
    Under Assumption \ref{assump:1}, for a fixed $\delta$, there exists $n_1$ such that $\forall n > n_1>n_0$ ($n_0$ is specified in Assumption \ref{assump:1}), $\forall a \geq 0$, it can be shown that 
    \begin{align*}
        \delta &\geq \Bigg|\sum_{\forall \boldsymbol{x}_i \in \{c \leq f(\boldsymbol{x}_i)\}\backslash x_+}\mathbb{P}\Bigg(G_n(\boldsymbol{x}_i)\geq -a-d_n(\boldsymbol{x}_i)\Bigg) +\\ 
        & \quad \sum_{\forall \boldsymbol{x}_i \in \{c > f(\boldsymbol{x}_i)\}\backslash x_{-}} \mathbb{P}\Bigg(G_n(\boldsymbol{x}_i)< a+|d_n(\boldsymbol{x}_i)|
        \Bigg) - (m-2)\Bigg|
    \end{align*}
    Then, 
    \begin{align*}
        & \mathbb{P}\Bigg(
        G_n(\boldsymbol{x}_+)\geq -a-d_n(\boldsymbol{x}_+)
        \bigcap 
        G_n(\boldsymbol{x}_-)< a+|d_n(\boldsymbol{x}_-)|
        \Bigg) - \delta
         \leq \\
        &\mathbb{P}\Bigg( 
            \mathrm{CS}_c^i \subseteq 
            \mathcal{X}_m(c)
            \subseteq 
            \mathrm{CS}_c^o
        \Bigg)\leq \\
        & \mathbb{P}\Bigg(
        G_n(\boldsymbol{x}_+)\geq -a-d_n(\boldsymbol{x}_+)
        \bigcap 
        G_n(\boldsymbol{x}_-)< a+|d_n(\boldsymbol{x}_-)|
        \Bigg).
    \end{align*}
    Furthermore, it can be shown that as $n\rightarrow \infty$, $\delta \rightarrow 0$,
    \begin{align*}
        \lim_{n\rightarrow \infty}\mathbb{P}\big( 
        \mathrm{CS}_c^i \subseteq 
        \mathcal{X}_m(c)
        \subseteq 
        \mathrm{CS}_c^o\big)=
        \lim_{n\rightarrow \infty}\mathbb{P}\big(
        G_n(\boldsymbol{x}_+)\geq -a-d_n(\boldsymbol{x}_+)
        \cap 
        G_n(\boldsymbol{x}_-)< a+|d_n(\boldsymbol{x}_-)|
        \big)=1.
    \end{align*}
\end{corollary}

Corollary \ref{corollary:finite_sharp_bound} states that when the training sample size is infinite and the closest point to the boundary is known, the containment statement of the constructed confidence sets is guaranteed. This is intuitive since, with an infinite training sample size, the estimated model $\hat{f}$ converges to the true model $f$, allowing us to easily distinguish points that are not on the boundary with the smallest threshold $a$ possible, as Corollary \ref{corollary:finite_sharp_bound} holds for every $a$. 

\section{Algorithms for constructing confidence sets}\label{sec:algorithm}
In this section, we describe an algorithm to construct confidence sets based on Theorem \ref{theorem:finite_main_bound}, labeled below as \hyperref[algorithm_confidence_set]{Construction Algorithm}. The validity of \hyperref[algorithm_confidence_set]{Construction Algorithm} does not require knowledge of the location of the boundary points and the training sample size can be finite. In addition, it has the ability to automatically adjust the number of points in the inflated boundary based on the training sample size and the structure of the model, making it readily applicable for real-world problems. Algorithms based on Theorem \ref{theorem:asy_main_bound} and Corollary \ref{corollary:finite_sharp_bound} are provided in the \hyperref[sec:appendix_start]{Appendix} to validate the corresponding theoretical results through simulation. They are not illustrated in the main text since these algorithms require knowledge of either boundary point positions or cannot guarantee type-I error rate control.

As a reference to compare with \hyperref[algorithm_confidence_set]{Construction Algorithm}, algorithms by inverting confidence intervals are described in the sub-section below.

Additional helper algorithms are defined in the \hyperref[sec:appendix_start]{Appendix}: Bootstrap Algorithm \hyperref[algorithm_boostrap]{$\mbox{BS}(\tilde{\boldsymbol{y}}, \tilde{\boldsymbol{X}}, \boldsymbol{X})$}, estimation of upper bound probability \hyperref[algorithm_est_upper_bound]{$\mbox{EstUpperBound}(a_m, \boldsymbol{G}_B, \hat{d}(\boldsymbol{x}_i), e_1, e_2)$} and estimation of lower bound probability \hyperref[algorithm_est_lower_bound]{$\mbox{EstLowerBound}(a_m, \boldsymbol{G}_B, \hat{d}(\boldsymbol{x}_i), e_1, e_2)$}.

\subsection{Confidence set construction algorithm}
In this section, we discuss the rationale for the \hyperref[algorithm_confidence_set]{Construction Algorithm}. Its goal is to find confidence sets with coverage probability whose lower bound $L$ according to Theorem \ref{theorem:finite_main_bound} is greater than a user-specified Target Lower Bound (TLB), say $1-\alpha$, and whose upper bound $U$ is as tight as possible. In other words, we want to find $e_1, e_2, a$ that yield the smallest value of $U-L$ with the constraint that $L\geq \text{TLB}$. In practice, we do not know the true upper bound $U$ nor the true lower bound $L$, so we define the Estimated Upper Bound (EUB) and the Estimated Lower Bound (ELB) as the bounds we can obtain and estimate. This algorithm requires no knowledge of ground truth values or locations, and can be directly implemented on real data.

Using Theorem \ref{theorem:asy_main_bound}, we know heuristically that as the training sample size increases, only the points closer to the level of interest matter. Therefore, it is rational to assume that for a fixed training sample size $n$, there exists an optimal value $e$ with $e_1=e, e_2=e$ that yields a threshold $a$ achieving the smallest value $U-L$ with the constraint that $L\geq \text{TLB}$. This is also supported by the fact that there exists values $e_1, e_2$ such that \eqref{eq:L.2} evaluates to 0 which achieves the smallest difference between \eqref{eq:U} and \eqref{eq:L.1}. When there is only one point on each side of the boundary level where \eqref{eq:L.2} evaluates to 0, then \eqref{eq:U} and \eqref{eq:L.1} coincide as shown in Corollary \ref{corollary:finite_sharp_bound}.

Combining the theorems and \hyperref[algorithm_confidence_set]{Construction Algorithm}, there is a duality between finding the locations of the uncertain prediction points and the tightness of the two bounds. Theorem \ref{theorem:asy_main_bound} indicates that when the model prediction is extremely accurate ($n\rightarrow \infty$), the only points where we are uncertain whether the prediction is greater than the level of interest are the points right on the boundary. However, when the sample is finite (the model accuracy in the test data is unknown), how do we find the location of the uncertain prediction points? Theorem \ref{theorem:finite_main_bound} and \hyperref[algorithm_confidence_set]{Construction Algorithm} show that by finding the tightest bound through varying the values of $e_1$ and $e_2$ is equivalent to locating the uncertain prediction points, which are the points in $e_1$ and $e_2$ where the tightest bound is achieved. This can be observed in Theorem \ref{theorem:finite_main_bound}, where the upper bound and lower bound difference achieve the minimum value when \ref{eq:L.1} includes the least amount of points (uncertain predictions), but \ref{eq:L.2} evaluates to 0 (points with high confidence that their values are greater than the level $c$). 

When constructing confidence sets for the realized outcome, the bootstrap portion of the \hyperref[algorithm_confidence_set]{Construction Algorithm} adds to the test point prediction a randomly selected residual (predicted value minus the realized outcome) generated from the non-selected samples during the model's training sample bootstrap step. Details are given in the algorithm \hyperref[algorithm_boostrap_observed_outcome]{$\mbox{BS}_e(\tilde{\boldsymbol{y}}, \tilde{\boldsymbol{X}}, \boldsymbol{X})$} in the \hyperref[sec:appendix_start]{Appendix}. As shown in the simulation below, this step takes account of the irreducible error and provides correct coverage.

\setalgname{Construction Algorithm}
\begin{algorithm*}
\caption{}\label{algorithm_confidence_set}
\begin{algorithmic}
\Require The targeted lower bound $TLB$, level of interest $c$, training data outcome $\tilde{\boldsymbol{y}}$ and design matrix $\tilde{\boldsymbol{X}}$. Test design matrix $\boldsymbol{X}$ with shape $(m, p)$. Objective. 
\If{Objective = Construction confidence set for expected outcome}:
    \State $\hat{\boldsymbol{y}}, std(\hat{\boldsymbol{y}}), \hat{\boldsymbol{Y}}_B \gets \text{BS}(\tilde{\boldsymbol{y}}, \tilde{\boldsymbol{X}}, \boldsymbol{X})$
\EndIf
\If{Objective = Construction confidence set for realized outcome}
    \State $\hat{\boldsymbol{y}}, std(\hat{\boldsymbol{y}}), \hat{\boldsymbol{Y}}_B \gets \text{BS}_e(\tilde{\boldsymbol{y}}, \tilde{\boldsymbol{X}}, \boldsymbol{X})$
\EndIf
\State $B$ by $m$ matrix $\boldsymbol{G}_B$ with $\boldsymbol{G}_B[b,:] \gets (\hat{\boldsymbol{Y}}_B[b,:] - \hat{\boldsymbol{y}})/std(\hat{\boldsymbol{y}})$
\State Let $\boldsymbol{x}_i = \boldsymbol{X}[i,:]$, and $i=1, ..., m, \hat{d}(\boldsymbol{x}_i) \gets \frac{\hat{\boldsymbol{y}}[i] - c}{std(\hat{\boldsymbol{y}})[i]}$
\State Sort the values $|\hat{d}(\boldsymbol{x}_i)|$ in ascending order and denote the resulting vector as $\hat{\boldsymbol{|d|}}_{sorted}$.
\State Initialize $r \gets Inf$
\State Initialize $a \gets 0$
\For{$e \in \hat{\boldsymbol{|d|}}_{sorted}$}
    \State $e_1, e_2 \gets e, e$
    \State Initialize $a_h, a_l \gets 10, 0.1$
    \State Initialize $ELB \gets 0$
    \While{$|ELB - TLB| > 0.001$}
        \State $a_m \gets (a_h+a_l)/2$
        \State $ELB \gets \mbox{EstLowerBound}(a_m, \boldsymbol{G}_B,\hat{d}(\boldsymbol{x}_i), e_1, e_2)$
        \If{$ELB > TLB$}:
            \State $a_h \gets a_m$
        \Else
            \State $a_l \gets a_m$
        \EndIf
    \EndWhile
    \State $EUB \gets \mbox{EstUpperBound}(a_m, \boldsymbol{G}_B,\hat{d}(\boldsymbol{x}_i), e_1, e_2)$
    \If{$r>EUB - ELB$}
        \State $a, r \gets a_m, EUB - ELB$
    \EndIf
\EndFor
\State Inner confidence set $\gets i:\hat{\boldsymbol{y}}[i] - a*std(\hat{\boldsymbol{y}})[i] \geq c$
\State Outer confidence set $\gets i:\hat{\boldsymbol{y}}[i] + a*std(\hat{\boldsymbol{y}})[i] \geq c$
\State \Return Inner confidence set, Outer confidence set, $ELB$, $EUB$
\end{algorithmic}

\end{algorithm*}

\subsection{Algorithm by inverting confidence intervals}
For comparison, we implemented two algorithms that use a point-wise confidence interval or simultaneous confidence interval to construct the confidence set. Given the upper and lower limits of point-wise or simultaneous confidence interval $\mathrm{CI}_u$ and $\mathrm{CI}_l$, then the inner and outer confidence sets can be constructed as:
\begin{align*}
    &\mathrm{CS}_c^i = \{\boldsymbol{x}_i: \mathrm{CI}_l(\boldsymbol{x}_i)\geq c\}\\
    &\mathrm{CS}_c^o = \{\boldsymbol{x}_i: \mathrm{CI}_u(\boldsymbol{x}_i)\geq c\}
\end{align*}
where $\mathrm{CI}_l(\boldsymbol{x}_i)$ is the lower bound and $\mathrm{CI}_u(\boldsymbol{x}_i)$ is the upper bound.

To construct the point-wise confidence intervals, for each test point, the algorithm simply calculates the ($1-\alpha/2$) and $\alpha/2$ quantiles of the bootstrap predictions, where $\alpha=1-TLB$. To construct the simultaneous confidence interval, the algorithm detailed in \cite{ren2024inverse} is implemented. 

When constructing confidence sets for the realized outcome, the algorithm adds to the prediction randomly selected residuals (predicted value minus the realized outcome) from the non-selected training samples during the bootstrap step. Then the confidence intervals are constructed for the modified prediction, similarly as above.

\section{Experimentation}\label{sec:experiment}
For the finite sample size theory and algorithms, we conducted a comprehensive assessment of the robustness of the constructed confidence sets considering various data types (continuous and binary) and model specifications (correctly and incorrectly specified). For the asymptotic theory and algorithm, we validate them under continuous data and correct model specification. For each simulation scenario, 500 simulation instances were performed. In each instance, 300 bootstrapped datasets were used to train 300 models, generating 300 sets of predictions for the corresponding test data containing 500 testing data points.

\subsection{Evaluation of finite sample size theory and algorithm}
We evaluate the validity of the \hyperref[algorithm_confidence_set]{Construction Algorithm} under two distinct scenarios: the model is correctly specified, or the model is misspecified but flexible enough to capture the underlying data structure. 

\subsubsection{Simulation with correct model structure specification}
To validate the algorithm when the assumptions are met, we generate linear and logistic regression data, and fit and predict the data with the same models structure as the data generation. The following models are used for data generation:
\begin{align}
    &(\mathbf{Linear}): y_i =  \sum_{j = 1}^{p}\beta_j x_{ij} + \epsilon_i \label{linear_eq}\\
    &(\mathbf{Logistic}): y_i \sim {\rm Bernoulli}(p_i),\quad  \log\left(\frac{p_i}{1-p_i} \right) =\sum_{j = 1}^{p}\beta_j x_{ij}  \label{logistic_eq},
\end{align}
where $i=1,...,n$ denote the index for independent observations and $n$ is the training sample size. For linear regression, the coefficients follow a standard normal distribution $\beta_j \iid  N(0, 1)$, generated for each simulation instance, and the irreducible error $\epsilon_i \iid N(0, \sigma)$. For logistic regression, $\beta_j \iid  {\rm Uniform}(1, 3)$ and $p_i$ denotes the probability of $y_i=1$. For both models, $x_{ij} \iid  {\rm Uniform}(-2, 2)$. We generate the test sample from the same underlying model and fix the size of the test sample to be $m= 500$. 

\begin{figure}
    \centering
    \includegraphics[scale = 0.6]{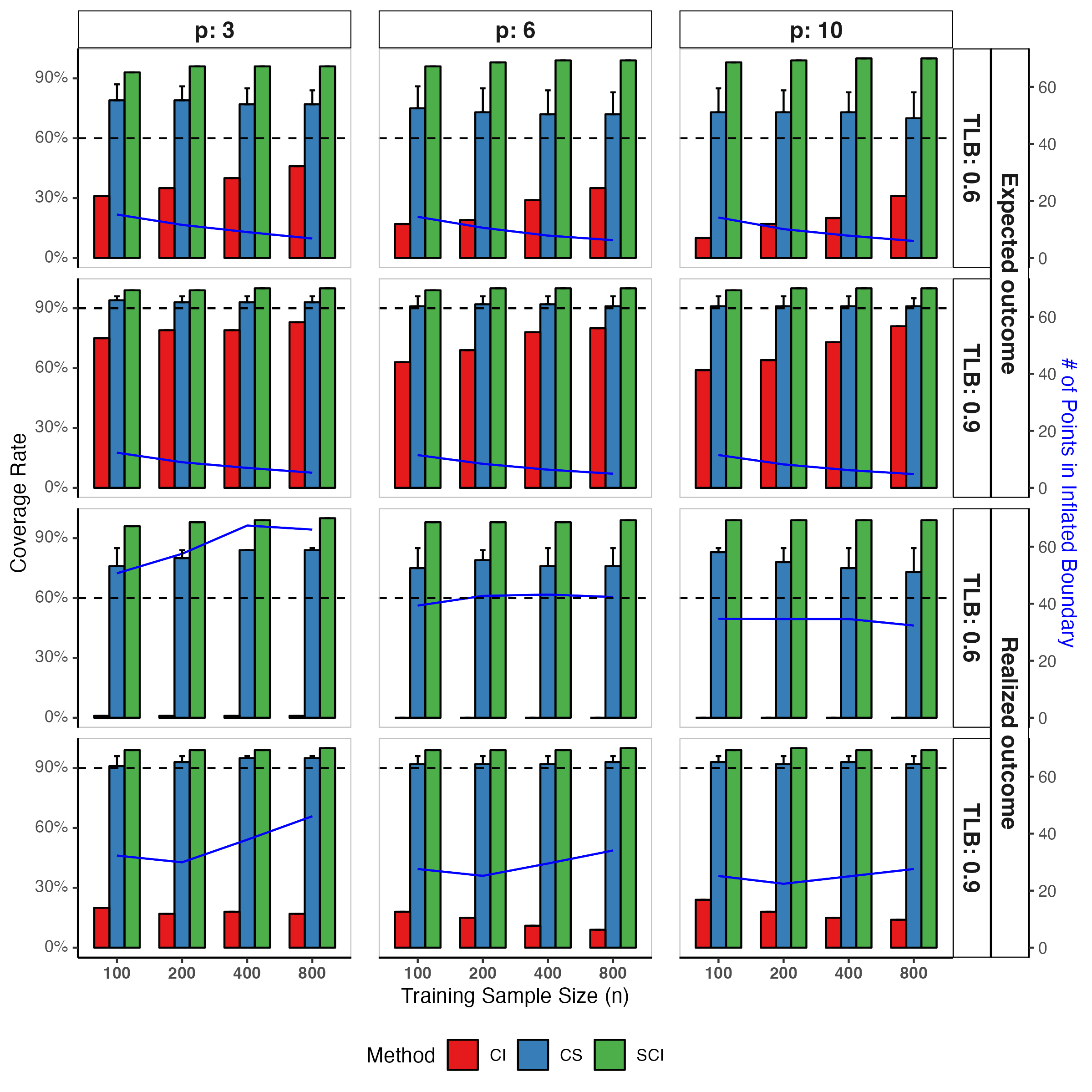}
    \caption{Validation of \hyperref[algorithm_confidence_set]{Construction Algorithm} under correct model specification with \textit{linear data}: confidence set (CS, blue), two naive algorithms by inverting point-wise confidence interval (CI, red) and inverting simultaneous confidence interval (SCI, green)—across different numbers of covariates $p$ (3, 6, and 10) and various training sample sizes $n$ (100, 200, 400, 800). The x-axis represents the training sample size, while the y-axis on the left indicates the coverage rate probability. The dashed horizontal lines represent the Targeted Lower Bound (TLB), with specific values of TLB = 0.6 and TLB = 0.9. The error bars displays the Estimated Upper Bound (EUB), which should be above the observed coverage rate. The blue line denotes the number of points in the inflated boundary for confidence set algorithm, corresponding to the right y-axis. The expected and realized outcomes are shown across the rows.}
    \label{fig:linear_sim_result}
\end{figure}

The first row of Figure \ref{fig:linear_sim_result} shows the coverage results of the confidence sets constructed for the {\it expected outcome} of the linear regression test set with number of covariates $p$ equal to 3, 6 and 10, and level of interest $c=0$. The coverage rate is maintained above the TLB and below EUB across the different training sample sizes. Furthermore, the algorithm becomes more confident by automatically finding the optimal number of points in the inflated boundary through finding the smallest range between the upper and lower bound. As the sample size increases, the algorithm shrinks the size of the inflated boundary, indicating that the algorithm captures the model uncertainty by becoming more confident as the model improves.

The second row of Figure \ref{fig:linear_sim_result} shows the coverage results of confidence set constructed for the {\it realized outcome} $y$ of the linear regression model with level of interest $c=0$. Similar to confidence set constructed for the expected outcome, the coverage rate is maintained between the TLB and EUB. The number of points in the inflated boundary does not decrease significantly with training sample size. We suspect that this is because the bulk part of the uncertainty is attributed to the irreducible error with a constant variance of 1 across different training sample sizes.

\begin{figure}
    \centering
    \includegraphics[scale = 0.6]{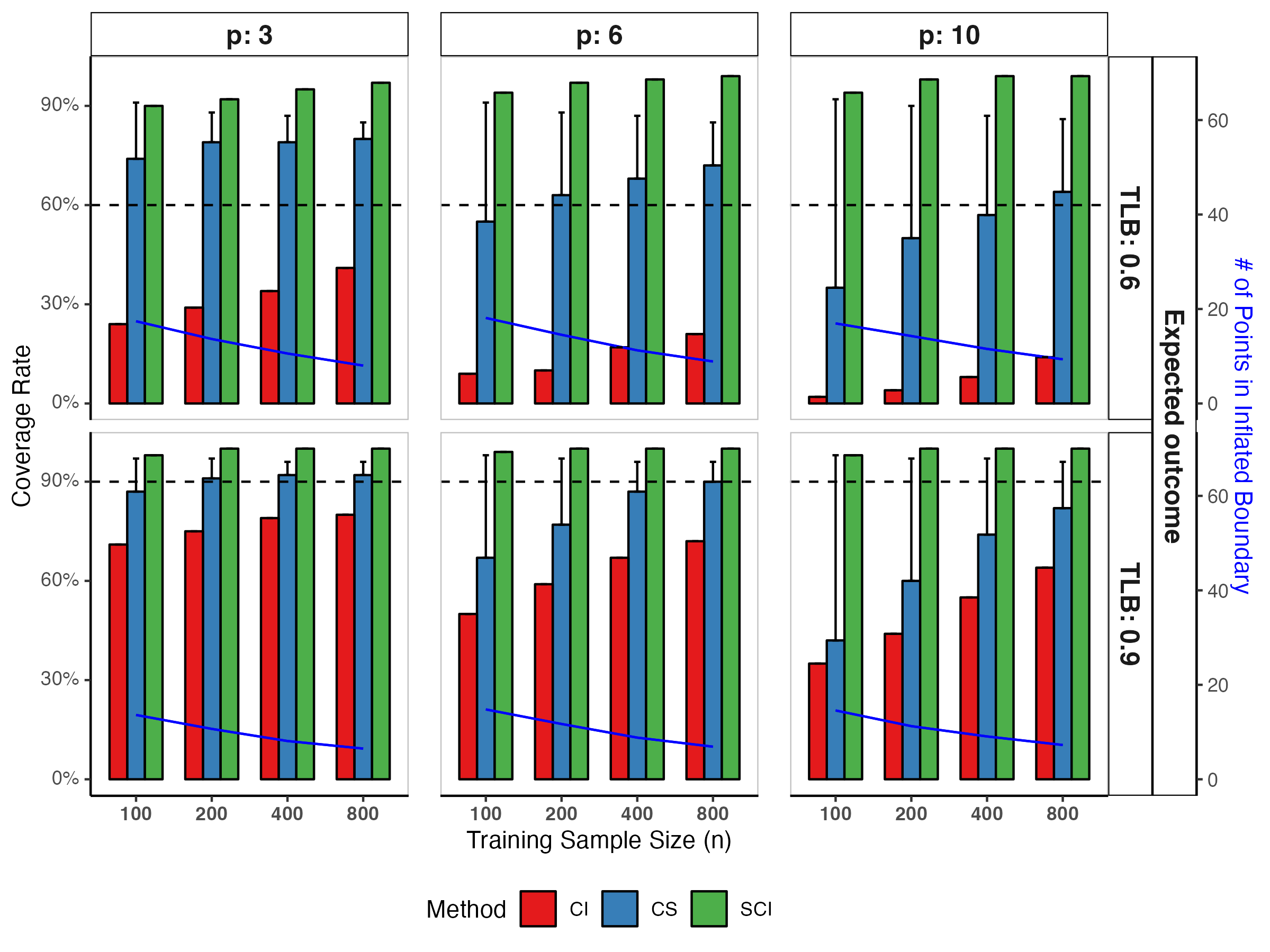}
    \caption{Validation of \hyperref[algorithm_confidence_set]{Construction Algorithm} under correct model specification with \textit{logistic data}: confidence set (CS, blue), two reference algorithms by inverting point-wise confidence interval (CI, red) and inverting simultaneous confidence interval (SCI, green)—across different numbers of covariates $p$ (3, 6, and 10) and various training sample sizes $n$ (100, 200, 400, 800). The x-axis represents the training sample size, while the y-axis on the left indicates the coverage rate probability. The dashed horizontal lines represent the Targeted Lower Bound (TLB), with specific values of TLB = 0.6 and TLB = 0.9. The error bars displays the Estimated Upper Bound (EUB), which should be above the observed coverage rate. The blue line denote the number of points in the inflated boundary for the confidence set algorithm, corresponding to the right y-axis. }
    \label{fig:logistic_sim_result}
\end{figure}

The experimentation results for confidence set constructed for the true class probability (expected outcome) are shown in Figure \ref{fig:logistic_sim_result} for number of covariates $p$ equal to 3, 6 and 10, and level of interest $c=0.5$. The coverage rates are maintained between the TLB and EUB similar to linear regression for $p=3$. However, as $p$ increases, the coverage rate decreases and drops below the TLB. This is because confidence sets are constructed for the true class probabilities which are not observed in our model fitting data (only the class labels 1 or 0 are observed). The model requires more training data to have an unbiased prediction, so Assumption \ref{assump:1} is violated. 

Compared to the method by inverting point-wise confidence intervals, the confidence set algorithm achieves significantly better coverage across every simulation scenario. Comparing to the naive method by inverting simultaneous confidence intervals, the confidence set algorithm achieves nominal coverage rate, whereas the simultaneous confidence intervals are more conservative (higher than nominal coverage rate) and yield lower power. 

\subsubsection{Simulation with misspecified model structure}
To simulate a scenario where the model structure is misspecified but capable of capturing the underlying data generating mechanism, we generate data from the following model:
\begin{align}
    &y_i = \beta_{0} + \beta_{1} \cos(\beta_{2}x_{i})
    + \epsilon_i \label{NN_eq}
\end{align}
where $i=1,...,n$ denotes the index for subjects and $n$ the training sample size. In this model, the coefficients are fixed at $\beta_{0}=1, \beta_{1} = 6, \beta_{2}=3$,  $x_{i} \iid  {\rm Uniform}(-2, 2)$, and the irreducible error is $\epsilon_i \iid N(0, 1)$. Unlike the linear model simulation, where the fitting model and data generating model coincide, the proposed model tries to estimate the complex relationship through the training sample. 

To evaluate a neural network model, the prediction model function in \hyperref[algorithm_confidence_set]{Construction Algorithm} is set as a fully-connected feedforward neural network with two hidden layers with 40 nodes in each hidden layer. The activation function for the hidden layers is the ReLu function. We split the training samples into two parts, with 80\% of the training sample serving as actual training set and 20\% of the training samples serving as validation set. We train the network with a learning rate of 0.01. After each epoch, we calculate the loss on the validation set. If the loss ceases to improve for 100 epochs, we stop the training and predict on the test set using the model with the lowest validation loss. The loss function is the mean square error loss. 

To evaluate a XGboost model \cite{chen2016xgboost}, we implemented a boosting model with 10 tree estimators that has a max depth of 6 and subsample proportion of 0.2 for the simulation. The level of interest is fixed at $0$ for both XGboost and neural network.

\begin{figure}
    \centering
    \includegraphics[scale = 0.6]{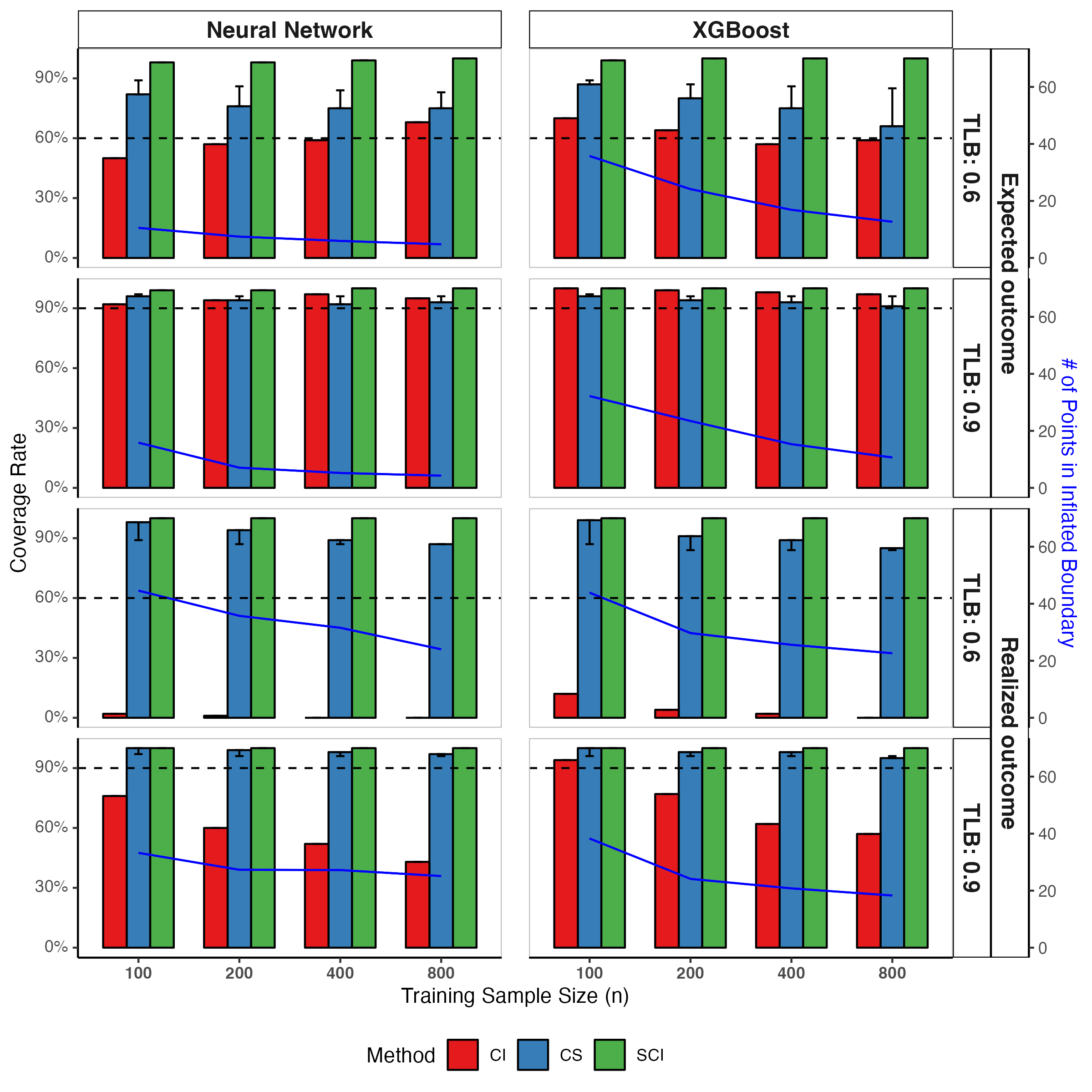}
    \caption{Validation of \hyperref[algorithm_confidence_set]{Construction Algorithm} under \textit{misspecified model with non-linear data}: confidence set (CS, blue), two naive algorithms by inverting point-wise confidence interval (CI, red) and inverting simultaneous confidence interval (SCI, green)—across different training sample sizes $n$ (100, 200, 400, 800). The x-axis represents the training sample size, while the y-axis on the left indicates the coverage rate probability. The dashed horizontal lines represent the Targeted Lower Bound (TLB), with specific values of TLB = 0.6 and TLB = 0.9. The error bars displays the Estimated Upper Bound (EUB), which should be above the observed coverage rate. The blue line denote the number of points in the inflated boundary for the confidence set algorithm, corresponding to the right y-axis. The result of neural network model and XGBoost model are shown across the columns. The expected and realized outcome are shown across the rows.}
    \label{fig:nonlinear_sim_result}
\end{figure}

Figure \ref{fig:demo_CS} shows one instance of the simulated data and the constructed confidence sets using the neural network model structure. The first row of Figure \ref{fig:nonlinear_sim_result} shows the coverage results of confidence set constructed for the {\it expected outcome}. Similar to the situation where the model structure is correctly specified, the coverage rate is maintained above the TLB and below the EUB across the different training sample sizes for both Neural network and XGBoost. Furthermore, the algorithm becomes more confident by automatically finding optimal number of points in the inflated boundary through finding the smallest range between the upper and lower bound. As the sample size increases, the algorithm shrinks the size of the inflated boundary,  indicating that the algorithm captures the model uncertainty by becoming more confident as the model improves. 

The second row of Figure \ref{fig:nonlinear_sim_result} shows the coverage results of confidence set constructed for the {\it realized outcome} $y$, which demonstrates a similar trend. Comparing to the naive algorithms using either CI or SCI, the confidence set algorithm maintained the coverage rate closest to the nominal level.

\subsection{Validation of asymptotic theory and algorithms}
\begin{figure}
    \centering
    \includegraphics[scale = 0.9]{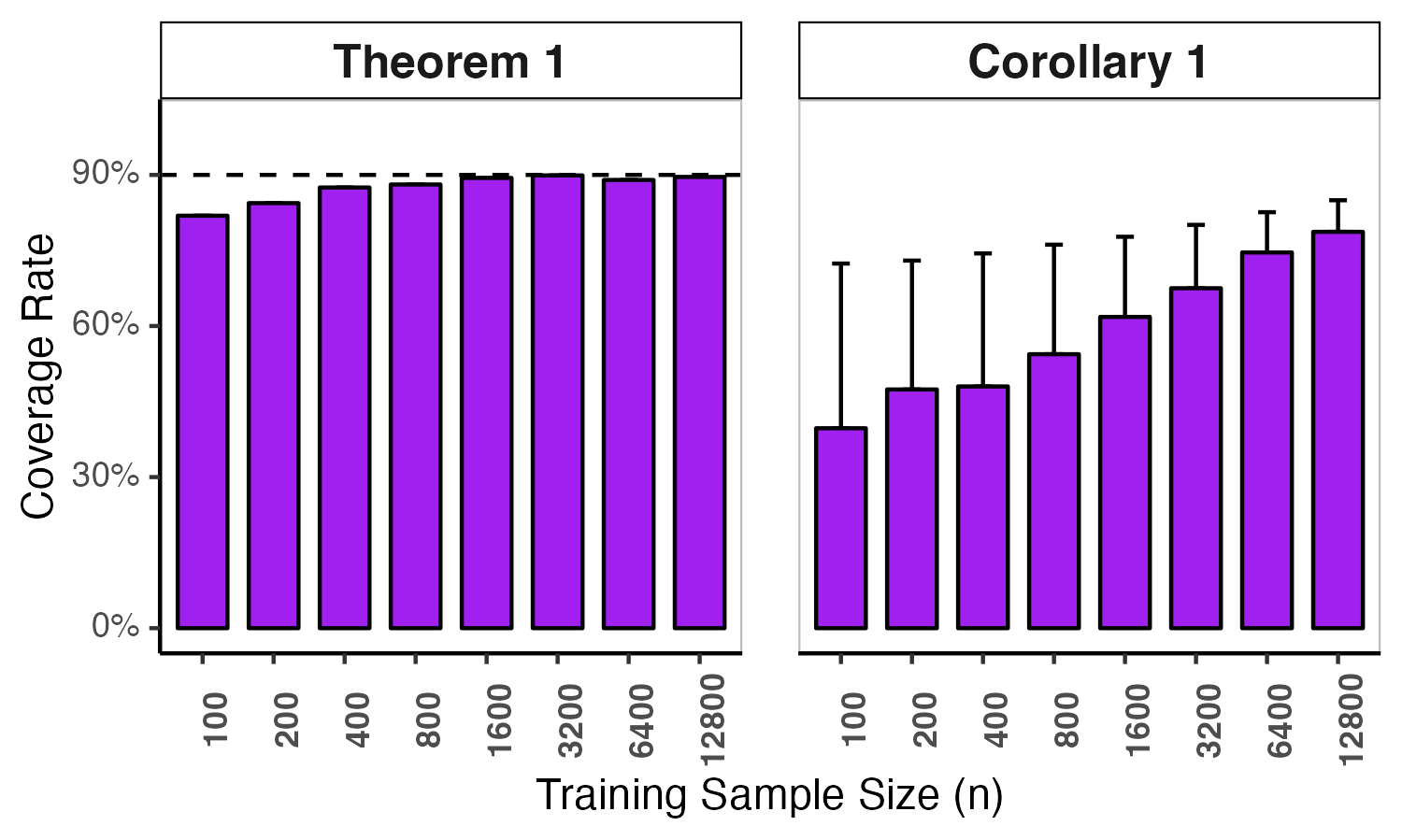}
    \caption{Validation of Theorem \ref{theorem:asy_main_bound} and Corollary \ref{corollary:finite_sharp_bound} under correct model specification with linear data: The x-axis represents the training sample size, while the y-axis on the left indicates the coverage rate probability. The dashed horizontal lines represent the Targeted Lower Bound (TLB), with specific values of TLB = 0.9. The error bars displays the Estimated Upper Bound (EUB), which should be above the observed coverage rate.}
    \label{fig:thm_coro_sim}
\end{figure}

To validate Theorem \ref{theorem:asy_main_bound} and Corollary \ref{corollary:finite_sharp_bound}, we generated continuous data using the linear model in Equation \ref{linear_eq}, assuming that the true locations of the test points $\boldsymbol{x}_i \in \mathcal{X}_m$ are known. The corresponding construction algorithms' details are in the \hyperref[sec:appendix_start]{Appendix}. The algorithms take in either the boundary points or the points closest to the boundary, which are unknown in practice but given in our simulation. For Theorem \ref{theorem:asy_main_bound}, the coverage rate converges to the target probability of 90\% as the sample size increases and remains stable after $n=1600$, as shown in the left plot of Figure \ref{fig:thm_coro_sim}. For Corollary \ref{corollary:finite_sharp_bound}, the coverage rate increases with the training sample size, while the difference between the mean estimated upper bound and the coverage rate decreases as the sample size grows, validating that $\delta$ shrinks as training sample size increases as shown in the Corollary \ref{corollary:finite_sharp_bound}. However, the convergence is slow in terms of the sample size if $a$ is small, making the Corollary less practical. 

\section{Data applications}\label{sec:application}
\subsection{Housing price prediction}
We used the data set "House Prices - Advanced Regression Techniques" \cite{lu2017hybrid} to demonstrate our algorithm for constructing confidence sets for a regression prediction model. After one-hot encoding and dropping features with over 50\% missing values, the dataset contained 1460 samples and 287 features. The goal is to identify which houses had a future sales price greater than the median housing price of \$157,024. The dataset was divided into 70\% training and 30\% testing sets (438 observations). We used random forest regression in scikit-learn \cite{buitinck2013api} as our prediction model. The mean absolute error on the test dataset was \$16,846.

Using the point prediction, we identified 219 points where the point predictions were greater than the median price. The sensitivity of this classification was 94.4\%, and the precision was 93.2\%. We then used our confidence set algorithm to construct the inner and outer sets for the expected and the realized outcome, with a type I error rate of 10\%. Using the inner confidence set as the positive classification, the precision increased to 95.7\% and 100.0\% for the sets constructed for the expected and the realized outcome, respectively. Using the outer confidence set as the positive classification, the sensitivity increased to 98.1\% and 100\% for the expected and the realized outcome, respectively. Table \ref{tab:house_pred_result} shows the number of points in different sets. There is a large discrepancy between the number of points in the confidence set constructed for the mean and the number of points for the confidence set constructed for the realized outcome, indicating that the prediction model captured only a small fraction of the information, and the irreducible noise was relatively large.

The result of confidence set constructed for the realized price demonstrates that the algorithm worked as intended, since the precision was 100\% for the inner set indicating that all inner points are true positive points. In addition, the sensitivity of the outer set points is 100\%, indicating that the outer set contains all the true positive points, which reflects the definition of the outer set. Based on the prespecified TLB of 90\% (type I error rate of 10\%), if we repeatedly sample the population and construct the confidence set, we would expect the inner set's precision and outer set's sensitivity to be 100\% for 90 instances out the 100 instances. 

\begin{table}[H]
    \centering
    \caption{Housing price prediction application. The table shows the number of samples in different confidence sets. The total number of evaluated tested samples is 438.}
    \begin{tabular}{lrr}
    \toprule
        Method &  Inner set points (precision) &  Outer set points (sensitivity) \\
    \midrule
    Point prediction for price &               219 (93.2\%) &               219 (94.4\%) \\
    Confidence set for expected price&               187 (95.7\%) &               260 (98.1\%) \\
    confidence set for realized price &                26 (100.0\%)&               438 (100.0\%) \\
    \bottomrule
    \end{tabular}
    \label{tab:house_pred_result}
\end{table}

\subsection{Time to sepsis prediction}
Sepsis, a life-threatening condition often seen in emergency departments, is a major contributor to mortality \cite{martischang2018promoting}. The importance of early sepsis detection cannot be overstated as it allows for the timely administration of antibiotics, which significantly improves clinical outcomes. Existing research demonstrates that for each hour of delayed treatment, there is an associated 4\%-8\% increase in mortality \cite{seymour2017time}. As a response to this, several machine learning models have been developed to predict sepsis before it occurs \cite{nemati2018interpretable, goh2021artificial}. However, optimizing limited resources in emergency departments necessitates more than just point predictions. It requires the identification of two distinct subsets of patients: those likely to develop sepsis within a certain time frame and those whose model predictions are uncertain. By doing so, physicians can allocate more resources to the high-risk subset.

Our algorithm aids in achieving this goal using the publicly available PhysioNet data \cite{reyna2019early}. To reduce computational time, we trained the algorithm using a randomly selected subset of the PhysioNet data, which comprises 600 unique patients admitted to emergency rooms. Of these 600 patients, 400 developed sepsis, and 200 did not, resulting in a total of 32,723 of observations (hourly) with 3,795 sepsis-positive time points across all patients. The test dataset includes 200 patients, 100 of whom developed sepsis, producing a total of 9,832 time points, with 960 being sepsis-positive. The data imputation and feature engineering process we followed is consistent with that outlined in \cite{yang2020explainable}. We implemented the XGboost model \cite{yang2020explainable}, with hyperparameters tuned on a validation set (20\%) within the training dataset.

This model predicts a binary label at each hour in the ICU, indicating whether the patient would develop sepsis in the subsequent six hours. The testing set consists of 9,832 time points, 960 of which were labeled as sepsis. 

The XGboost model predicts logit values for each time point, utilizing a cutoff of 0.0 on the logit scale and 0.5 on the probability scale to distinguish between the two classes. This means that the true set will include only patients with at least 50\% of probability sepsis, resulting in on average half of patients in the set will be diagnosed with sepsis. 

We implemented our confidence set algorithm with the lower bound of the Type I error rate set at 5\%. The point predictions identified 411 time points where the patient would likely develop sepsis within six hours. In contrast, the inner confidence set captured seven time points, at which patients would \textit{all} cross the decision boundary and be classified as sepsis with at least 95\% confidence. As illustrated in Table \ref{tab:sepsis_pred_result}, the point prediction achieved 21\% precision, whereas the inner set reached 57\% precision. This indicates that 57\% of the patients in the inner set are diagnosed with sepsis, demonstrating that the algorithm works as intended. 

Furthermore, the point prediction captured just 9\% of the true positives, indicating a sensitivity of 9\%. The outer confidence set, after excluding the inner set, contained 3051 time points about which the model was uncertain. By classifying this subset as potential sepsis cases, we achieved a sensitivity of 52.7\%. This indicates that outer set contains 52.7\% of all the patients diagnosed with sepsis. 

Using only the point prediction would cause doctors not only to focus more on patients who would not develop sepsis but also to risk overlooking patients who genuinely developed sepsis. With the supplementary information from the confidence sets, doctors can discern the areas where the model's predictions are rather uncertain. Specifically, the physician should concentrate on the seven time points in the inner set, while also paying attention to the 3051 time points in the outer set, which are quite likely to be classified as positive if the model is fitted on a resampled training set from the population.

\begin{table}[H]
    \centering
    \caption{Sepsis onset-time prediction problem. The table shows the number of samples in different confidence sets. The total number of evaluated tested samples is 6711.}
    \begin{tabular}{lrr}
    \toprule
        Method &  Inner set points (precision) &  Outer set points (sensitivity) \\
    \midrule
    Point prediction for sepsis probability &               411 (21.4\%) &               411 (9.2\%) \\
    Confidence set for sepsis probability &               7 (57.1\%) &               3051 (52.7\%)\\
    \bottomrule
    \end{tabular}
    \label{tab:sepsis_pred_result}
\end{table}

\section{Discussion}\label{sec:discussion}
In this paper, we presented a model-agnostic framework for uncertainty quantification and interpretation in continuous and binary outcome prediction models. The goal is to identify the subset of samples that are greater or less than a certain value of interest. We developed algorithms to construct an inner set that achieves 100\% precision and an outer set that achieves 100\% sensitivity, with a predefined probability for classifying which points are greater than or less than a certain value. This additional information provided by the inner and outer sets is valuable in practice for making decisions based on the predictions. For instance, in emergency rooms, doctors can allocate more resources to the subset of patients with a high probability that their predicted hours until sepsis occurrence is less than a certain threshold. This can improve patient outcomes and increase the efficiency of healthcare resource utilization.

The assumptions for the theorems we presented require that the prediction model is unbiased for the unseen test data. This is a relatively weak assumption when using neural networks trained on a large amount of data \cite{lu2021learning}. However, when there is not enough data or there is a distributional shift in the test dataset \cite{malinin2shifts}, the unbiasedness assumption may not hold. Additionally, although the theorems we presented hold for finite sample sizes, the validity of the bootstrap algorithm hinges on the asymptotic sample size \cite{efron1992bootstrap}. Thus, in practice, for precise control of the type-I error rate, the number of training samples must be sufficiently large, depending on the complexity of the data and the true generating distribution. However, with data collection technology advancing at a high speed, the sample size for most practical use cases should be sufficient for good performance of the bootstrap algorithm.

Several issues deserve further investigation. Binary classification problems arise more often in real-world applications than regression problems, where the level of interest for the label probability is often vague. Thus, more work is required to develop strategies for utilizing confidence sets on binary outcomes. Moreover, the current method controls the family-wise error rate, but many problems may not require such strong control. An algorithm for controlling the false-discovery rate awaits further development.

\bibliographystyle{plainnat}
\bibliography{mybib}

\newpage
\appendix

\setcounter{algorithm}{0}
\section{Appendix}\label{sec:appendix_start}

\subsection{Asymptotic sample size algorithms}
\hyperref[algorithm_confidence_set_theorem1]{Theorem \ref{theorem:asy_main_bound} Based Algorithm}, designed for large sample sizes, requires knowledge of the boundary point positions and existence. \hyperref[algorithm_confidence_set_corollary1]{Corollary \ref{corollary:finite_sharp_bound} Based Algorithm}, can only estimate the upper probability bound, without actual control of the lower bound. These prerequisites and limitations significantly restrict the practical applicability of the asymptotic algorithms. Although their real-world utility is constrained, we provide a brief overview of these algorithms as they are employed in simulations to validate Theorem \ref{theorem:asy_main_bound} and Corollary \ref{corollary:finite_sharp_bound}.

To construct confidence sets based on Theorem \ref{theorem:asy_main_bound}, if $\boldsymbol{x}_i \in \{d_n(\boldsymbol{x}_i)=0\}$ are known, for each set of bootstrapped $G_n(\boldsymbol{x}_i)$, we find the minimum of $G_n(\boldsymbol{x}_i)$ for $\forall \boldsymbol{x}_i \in \{d_n(\boldsymbol{x}_i)=0\}$, and we take the proportion of bootstrap samples for which the minimum is greater than $-a$ as the estimate containment probability. We gradually increase the $a$ and repeat the steps above until the containment probability achieves the targeted probability bound.

\setalgname{Theorem \ref{theorem:asy_main_bound} Based Algorithm}
\begin{algorithm}
\caption{}
\begin{algorithmic}
\Require The targeted lower bound $TLB$, level of interest $c$, training data outcome $\tilde{\boldsymbol{y}}$ and design matrix $\tilde{\boldsymbol{X}}$. Test design matrix $\boldsymbol{X}$ with shape $(m, p)$, boundary points $\{\boldsymbol{x}_i: d_n(\boldsymbol{x}_i)=0\}$, $\eta=0.001$.
\State $\hat{\boldsymbol{y}}, std(\hat{\boldsymbol{y}}), \hat{\boldsymbol{Y}}_B \gets \text{BS}(\tilde{\boldsymbol{y}}, \tilde{\boldsymbol{X}}, \boldsymbol{X})$
\State $B$ by $m$ matrix $\boldsymbol{G}_B$ with $\boldsymbol{G}_B[b,:] \gets (\hat{\boldsymbol{Y}}_B[b,:] - \hat{\boldsymbol{y}})/std(\hat{\boldsymbol{y}})$
\State Initialize $ELB \gets Inf$
\State Initialize $a \gets \eta$
\While{$|ELB - TLB| > \eta$}
    \State $ELB \gets \frac{1}{B}\sum_{b=1}^B \mathbb{I}\left(\argmin_{\{\boldsymbol{x}_i: d_n(\boldsymbol{x}_i)=0\}}\boldsymbol{G}_B[b,:]>-a\right) $
    \State $a \gets a+\eta$
\EndWhile
\State Inner confidence set $\gets i:\hat{\boldsymbol{y}}[i] - a*std(\hat{\boldsymbol{y}})[i] \geq c$
\State Outer confidence set $\gets i:\hat{\boldsymbol{y}}[i] + a*std(\hat{\boldsymbol{y}})[i] \geq c$
\State \Return Inner confidence set, Outer confidence set, $ELB$
\end{algorithmic}
\label{algorithm_confidence_set_theorem1}
\end{algorithm}

Based on Corollary \ref{corollary:finite_sharp_bound}, using the position information of the two closest point to the boundary and fixing $a=1$, the second algorithm takes the proportion of the times that, $G_n(\boldsymbol{x}_+)\geq -a-d_n(\boldsymbol{x}_+)\bigcap G_n(\boldsymbol{x}_-)< a+|d_n(\boldsymbol{x}_-)|$ holds for the bootstrapped samples, as upper bound.

\setalgname{Corollary \ref{corollary:finite_sharp_bound} Based Algorithm}
\begin{algorithm}
\caption{}
\begin{algorithmic}
\Require The targeted lower bound $TLB$, level of interest $c$, training data outcome $\tilde{\boldsymbol{y}}$ and design matrix $\tilde{\boldsymbol{X}}$. Test design matrix $\boldsymbol{X}$ with shape $(m, p)$, two closest point to the boundary $\{\boldsymbol{x}_+, \boldsymbol{x}_-\}$.
\State $\hat{\boldsymbol{y}}, std(\hat{\boldsymbol{y}}), \hat{\boldsymbol{Y}}_B \gets \text{BS}(\tilde{\boldsymbol{y}}, \tilde{\boldsymbol{X}}, \boldsymbol{X})$
\State $B$ by $m$ matrix $\boldsymbol{G}_B$ with $\boldsymbol{G}_B[b,:] \gets (\hat{\boldsymbol{Y}}_B[b,:] - \hat{\boldsymbol{y}})/std(\hat{\boldsymbol{y}})$
\State Initialize $a \gets 1$
\State $EUB \gets \frac{1}{B}\sum_{b=1}^B \mathbb{I}\left(G_n(\boldsymbol{x}_+)\geq -a-d_n(\boldsymbol{x}_+)\bigcap G_n(\boldsymbol{x}_-)< a+|d_n(\boldsymbol{x}_-)|\right)$
\State Inner confidence set $\gets i:\hat{\boldsymbol{y}}[i] - a*std(\hat{\boldsymbol{y}})[i] \geq c$
\State Outer confidence set $\gets i:\hat{\boldsymbol{y}}[i] + a*std(\hat{\boldsymbol{y}})[i] \geq c$
\State \Return Inner confidence set, Outer confidence set, $EUB$
\end{algorithmic}
\label{algorithm_confidence_set_corollary1}
\end{algorithm}

\subsection{Helper algorithms}

\setalgname{$\mbox{BS}(\tilde{\boldsymbol{y}}, \tilde{\boldsymbol{X}}, \boldsymbol{X})$}
\begin{algorithm}[H]
\begin{algorithmic}[1]
\caption{Bootstrap Algorithm}
\Require Training data outcome $\tilde{\boldsymbol{y}}$ and design matrix $\tilde{\boldsymbol{X}}$. Test design matrix $\boldsymbol{X}$ with shape $(m, p)$.
\Require Number of Bootstraps $B$ and a fixed model function $f(\boldsymbol{\beta}, \boldsymbol{X})$ with parameters $\boldsymbol{\beta}$, and a loss function $L(f(\boldsymbol{\beta}, \boldsymbol{X}), y)$.
\State Initialize an empty $B \times m$ matrix $\hat{\boldsymbol{Y}}_B$.
\State $\hat{\boldsymbol{\beta}} \gets \argmin_{\boldsymbol{\beta}} L(f(\boldsymbol{\beta}, \tilde{\boldsymbol{X}}), \tilde{\boldsymbol{y}})$
\State $\hat{\boldsymbol{y}} \gets f(\hat{\boldsymbol{\beta}}, \boldsymbol{X})$
\For{$b$ in $1,..,B$}
    \State $\tilde{\boldsymbol{y}}_b,\tilde{\boldsymbol{X}}_b \gets $ sample with replacement on $\tilde{\boldsymbol{y}}, \tilde{\boldsymbol{X}}$
    \State $\hat{\boldsymbol{\beta}}_b \gets \argmin_{\boldsymbol{\beta}} L(f(\boldsymbol{\beta}, \tilde{\boldsymbol{X}}_b), \tilde{\boldsymbol{y}}_b)$
    \State $\hat{\boldsymbol{Y}}_B[b,:] \gets f(\hat{\boldsymbol{\beta}}_b, \boldsymbol{X})$
\EndFor
\State $std(\hat{\boldsymbol{y}}) \gets$   column-wise standard deviation of the matrix $\hat{\boldsymbol{Y}}_B$
\State \Return $\hat{\boldsymbol{y}}, std(\hat{\boldsymbol{y}}), \hat{\boldsymbol{Y}}_B$
\end{algorithmic}
\label{algorithm_boostrap}
\end{algorithm}

\setalgname{$\mbox{BS}_e(\tilde{\boldsymbol{y}}, \tilde{\boldsymbol{X}}, \boldsymbol{X})$}
\begin{algorithm}[H]
\begin{algorithmic}[1]
\caption{Bootstrap Algorithm for realized outcome}
\Require Training data outcome $\tilde{\boldsymbol{y}}$ and design matrix $\tilde{\boldsymbol{X}}$. Test design matrix $\boldsymbol{X}$ with shape $(m, p)$.
\Require Number of Bootstraps $B$ and a fixed model function $f(\boldsymbol{\beta}, \boldsymbol{X})$ with parameters $\boldsymbol{\beta}$, and a loss function $L(f(\boldsymbol{\beta}, \boldsymbol{X}), y)$.
\State Initialize an empty $B \times m$ matrix $\hat{\boldsymbol{Y}}_B$.
\State $\hat{\boldsymbol{\beta}} \gets \argmin_{\boldsymbol{\beta}} L(f(\boldsymbol{\beta}, \tilde{\boldsymbol{X}}), \tilde{\boldsymbol{y}})$
\State $\hat{\boldsymbol{y}} \gets f(\hat{\boldsymbol{\beta}}, \boldsymbol{X})$
\For{$b$ in $1,..,B$}
    \State $\tilde{\boldsymbol{y}}_b,\tilde{\boldsymbol{X}}_b \gets $ sample with replacement on $\tilde{\boldsymbol{y}}, \tilde{\boldsymbol{X}}$
    \State $\tilde{\boldsymbol{y}}_{-b},\tilde{\boldsymbol{X}}_{-b}$ are samples not selected during the current bootstrap.
    \State $\hat{\boldsymbol{\beta}}_b \gets \argmin_{\boldsymbol{\beta}} L(f(\boldsymbol{\beta}, \tilde{\boldsymbol{X}}_b), \tilde{\boldsymbol{y}}_b)$
    \State $e_{-b}\gets \tilde{\boldsymbol{y}}_{-b}-f(\hat{\boldsymbol{\beta}}_b, \tilde{\boldsymbol{X}}_{-b})$
    \State Randomly select from $e_{-b}$ to match the dimension of the current bootstrap: $e_b$
    \State $\hat{\boldsymbol{Y}}_B[b,:] \gets f(\hat{\boldsymbol{\beta}}_b, \boldsymbol{X})+e_b$
\EndFor
\State $std(\hat{\boldsymbol{y}}) \gets$   column-wise standard deviation of the matrix $\hat{\boldsymbol{Y}}_B$
\State \Return $\hat{\boldsymbol{y}}, std(\hat{\boldsymbol{y}}), \hat{\boldsymbol{Y}}_B$
\end{algorithmic}
\label{algorithm_boostrap_observed_outcome}
\end{algorithm}

\setalgname{$\mbox{EstUpperBound}(a_m, \boldsymbol{G}_B, \hat{d}(\boldsymbol{x}_i), e_1, e_2)$}
\begin{algorithm}[H]
\begin{algorithmic}[1]
\caption{Estimation function for the uppper bound as shown in Equation \ref{eq:U}}
\Require Threshold $a_m$, $B$ by $m$ matrix $\boldsymbol{G}_B$, a vector of distance $\hat{d}(\boldsymbol{x}_i)$, $e_1$, $e_2$
\State 
\begin{align*}
    p &\gets \\
    &\frac{1}{B} \sum_{b=1}^{B} \mathbb{I}\bigg(
    \min_{\boldsymbol{x}_i \in \{0 \leq \hat{d}(\boldsymbol{x}_i)\leq e_1\}}|\boldsymbol{G}_B[b,i]|<a_m+\max_{\boldsymbol{x}_i \in \{0 \leq \hat{d}(\boldsymbol{x}_i)\leq e_1\}}|\hat{d}(\boldsymbol{x}_i)| \bigg) \times\\
    &\quad \quad \quad\mathbb{I}\bigg(\max_{\boldsymbol{x}_i \in \{-e_2 \leq \hat{d}(\boldsymbol{x}_i)< 0\}}|\boldsymbol{G}_B[b,i]|<a_m+\max_{\boldsymbol{x}_i \in \{-e_2 \leq \hat{d}(\boldsymbol{x}_i)< 0\}}|\hat{d}(\boldsymbol{x}_i)|
    \bigg)
\end{align*}
\State \Return $p$
\end{algorithmic}
\label{algorithm_est_upper_bound}
\end{algorithm}

\setalgname{$\mbox{EstLowerBound}(a_m, \boldsymbol{G}_B, \hat{d}(\boldsymbol{x}_i), e_1, e_2)$}
\begin{algorithm}[H]
\begin{algorithmic}[1]
\caption{Estimation function for the lower bound as shown in Equation \ref{eq:L.1}+\ref{eq:L.2}}
\Require Threshold $a_m$, $B$ by $m$ matrix $\boldsymbol{G}_B$, a vector of distance $\hat{d}(\boldsymbol{x}_i)$, $e_1$, $e_2$
\State 
\begin{align*}
    p &\gets \\
    &\frac{1}{B} \sum_{b=1}^{B} \mathbb{I}\left(\max_{\boldsymbol{x}_i \in \{-e_2 \leq \hat{d}(\boldsymbol{x}_i)< e_1\}}|\boldsymbol{G}_B[b,i]|<a_m+\min_{\boldsymbol{x}_i \in \{-e_2 \leq \hat{d}(\boldsymbol{x}_i)< e_1\}}|\hat{d}(\boldsymbol{x}_i)|\right)+\\
    &\sum_{\boldsymbol{x}_i \in \{\hat{d}(\boldsymbol{x}_i)> e_1\}}\frac{1}{B}\sum_{b=1}^{B} \mathbb{I}\left(\boldsymbol{G}_B[b,i] \geq -a_m-\hat{d}(\boldsymbol{x}_i) \right)+\\
    &\sum_{\boldsymbol{x}_i \in \{\hat{d}(\boldsymbol{x}_i)< -e_2\}}\frac{1}{B}\sum_{b=1}^{B} \mathbb{I}\left(\boldsymbol{G}_B[b,i] \geq a_m+|\hat{d}(\boldsymbol{x}_i)| \right)-\\
    &\mathrm{CARD}\Bigg\{\boldsymbol{x}_i: \hat{d}(\boldsymbol{x}_i)> e_1\Bigg\}-\mathrm{CARD}\Bigg\{\boldsymbol{x}_i: \hat{d}(\boldsymbol{x}_i)< e_2\Bigg\}
\end{align*}
\State \Return $p$
\end{algorithmic}
\label{algorithm_est_lower_bound}
\end{algorithm}

\subsection{Proofs}
\begin{lemma}\label{lemma:finite_equi}
Define the following events as:
    \begin{align*}
        E_1 &= 
        \mathrm{CS}_c^i \subseteq 
            \mathcal{X}_m(c)
            \subseteq 
            \mathrm{CS}_c^o\\
        &=
        \left\{\boldsymbol{x}_i \in  \mathcal{X}_{m}:\frac{\hat f_n(\boldsymbol{x}_i )-c}{\tau_n\sigma(\boldsymbol{x}_i )}\geq a\right\} \subseteq 
        \{\boldsymbol{x}_i \in  \mathcal{X}_{m}:f(\boldsymbol{x}_i )\geq c\} 
        \subseteq 
        \left\{\boldsymbol{x}_i \in  \mathcal{X}_{m}:\frac{\hat f_n(\boldsymbol{x}_i )-c}{\tau_n\sigma(\boldsymbol{x}_i )}\geq -a\right\}\\
        E_2 &=  \left\{ \forall \boldsymbol{x}_i  \in \{d_n(\boldsymbol{x}_i )\geq 0\}:\frac{\hat f_n(\boldsymbol{x}_i )-c}{\tau_n\sigma(\boldsymbol{x}_i )}\geq -a\right\}
            \bigcap 
            \left\{\forall \boldsymbol{x}_i \in \{d_n(\boldsymbol{x}_i )<0\}:\frac{\hat f_n(\boldsymbol{x}_i )-c}{\tau_n\sigma(\boldsymbol{x}_i )}< a  \right\}.
    \end{align*}
Then, $E_1$ and $E_2$ are equivalent.
\end{lemma}
\begin{proof}
    Observe that the following sets are equal:
    \begin{align*}
        &\{d_n(\boldsymbol{x}_i )\geq 0\} = \{f(\boldsymbol{x}_i)\geq c \}\\
        &\{d_n(\boldsymbol{x}_i )< 0\} = \{f(\boldsymbol{x}_i)< c \}.
    \end{align*}
    First, we show that the event $E_2$ implies $E_1$,
    $\forall \boldsymbol{x}_i \in \{f(\boldsymbol{x}_i)\geq c \}$, directly from the first sub-event in $E_2$ we have,
    $$\frac{\hat f_n(\boldsymbol{x}_i)-c}{\tau_n\sigma(\boldsymbol{x}_i)}\geq -a$$
    Therefore, $\{f(\boldsymbol{x}_i)\geq c\} \subseteq \left\{\frac{\hat f_n(\boldsymbol{x}_i)-c}{\tau_n\sigma(\boldsymbol{x}_i)}\geq -a\right\}$ holds.
    
    Then, we show $\left\{\frac{\hat f_n(\boldsymbol{x}_i)-c}{\tau_n\sigma(\boldsymbol{x}_i)}\geq a\right\} \subseteq 
    \{f(\boldsymbol{x}_i)\geq c\} $ which is equivalent to showing
    $\left\{\frac{\hat f_n(\boldsymbol{x}_i)-c}{\tau_n\sigma(\boldsymbol{x}_i)}< a\right\} \supset 
    \{f(\boldsymbol{x}_i)< c\} $ which directly follows from the second sub-event in $E_2$.
    
    Second, let's show $E_1$ implies $E_2$:
    If $\exists \boldsymbol{x}_i \in \{f(\boldsymbol{x}_i) \geq c\}, \frac{\hat f_n(\boldsymbol{x}_i)-c}{\tau_n\sigma(\boldsymbol{x}_i)}<-a $ then $\{f(\boldsymbol{x}_i) \geq c\}\not \subseteq \left\{\frac{\hat f_n(\boldsymbol{x}_i)-c}{\tau_n\sigma(\boldsymbol{x}_i)}\geq-a \right\}$, contradiction to $E_1$.
    If $\exists \boldsymbol{x}_i \in \{f(\boldsymbol{x}_i) < c\}, \frac{\hat f_n(\boldsymbol{x}_i)-c}{\tau_n\sigma(\boldsymbol{x}_i)}\geq a $ then $\left\{\frac{\hat f_n(\boldsymbol{x}_i)-c}{\tau_n\sigma(\boldsymbol{x}_i)}\geq a \right\} \not \subseteq \{f(\boldsymbol{x}_i) \geq c\} $, contradiction to $E_1$.
\end{proof}

From Lemma \ref{lemma:finite_equi}, we can see if we can obtain $a$ such that 
\begin{align*}
    \mathbb{P}\left(\inf_{\boldsymbol{x}_i \in \{d_n(\boldsymbol{x}_i)\geq 0\}}\frac{\hat f_n(\boldsymbol{x}_i)-c}{\tau_n\sigma(\boldsymbol{x}_i)}\geq -a \bigcap \sup_{\boldsymbol{x}_i \in \{d_n(\boldsymbol{x}_i)<0\}}\frac{\hat f_n(\boldsymbol{x}_i)-c}{\tau_n\sigma(\boldsymbol{x}_i)}< a\right)=1-\alpha,
\end{align*}
we would have control over the type I error rate of $E_1$. The obstruction is that this probability cannot be obtained without strong assumptions on the estimator $\hat f(\boldsymbol{x}_i)$ in practice. On the other hand, it would be much easier to obtain probabilities that involve $\frac{\hat f_n(\boldsymbol{x}_i)-\mu(\boldsymbol{x}_i)}{\tau_n\sigma(\boldsymbol{x}_i)}$ contrary to $\frac{\hat f_n(\boldsymbol{x}_i)-c}{\tau_n\sigma(\boldsymbol{x}_i)}$.

\begin{lemma}\label{lemma:intersection_bound}
    For all sets $A$ and $B$, we have that
    $$
    \mathbb{P}(A \cap B) \geq \mathbb{P}(A) + \mathbb{P}(B) - 1
    $$
\end{lemma}
\begin{proof}
    Let $A^c$ denote the complement of $A$ then
    $$\mathbb{P}(A \cap B) = 1 - \mathbb{P}(A^c \cup B^c) \geq 1 - \mathbb{P}(A^c) - \mathbb{P}(B^c) = \mathbb{P}(A)+\mathbb{P}(B) - 1   $$
\end{proof}

\begin{lemma}\label{lemma:main_bound}
    Condition on that the sets $\{d_n(\boldsymbol{x}_i)=0\}$, $\{d_n(\boldsymbol{x}_i)>0\}$ and $\{d_n(\boldsymbol{x}_i)<0\}$ are non-empty, then
    \begin{align*}
        & \mathbb{P}\left(
            \inf_{\boldsymbol{x}_i \in \{d_n(\boldsymbol{x}_i)=0\}}G_n(\boldsymbol{x}_i)\geq -a\right)\geq \mathbb{P}\big( \mathrm{CS}_c^i \subseteq 
            \mathcal{X}_m(c)
            \subseteq 
            \mathrm{CS}_c^o
        \big)\geq \\
        & \mathbb{P}\left(
            \inf_{\boldsymbol{x}_i \in \{d_n(\boldsymbol{x}_i)=0\}}G_n(\boldsymbol{x}_i)\geq -a\right)
            +
            \mathbb{P}\left(\inf_{\boldsymbol{x}_i \in \{d_n(\boldsymbol{x}_i)>0\}}G_n(\boldsymbol{x}_i)\geq -a-\inf_{\boldsymbol{x}_i \in \{d_n(\boldsymbol{x}_i)>0\}}d_n(\boldsymbol{x}_i)\right)+
            \\
            &\quad \mathbb{P}\left(\sup_{\boldsymbol{x}_i \in \{d_n(\boldsymbol{x}_i)<0\}}G_n(\boldsymbol{x}_i)< a+\inf_{\boldsymbol{x}_i \in \{d_n(\boldsymbol{x}_i)<0\}}|d_n(\boldsymbol{x}_i)|
        \right)-2
    \end{align*}
    Note that both $\inf_{\boldsymbol{x}_i \in \{d_n(\boldsymbol{x}_i)>0\}}d_n(\boldsymbol{x}_i)$ and $\inf_{\boldsymbol{x}_i \in \{d_n(\boldsymbol{x}_i)<0\}}|d_n(\boldsymbol{x}_i)|$ are constant conditioned on $\mathcal{X}_m$.
\end{lemma}

\begin{proof}
    Using Lemma \ref{lemma:finite_equi}, we obtain the first inequality by $E_1$ implies a sub-event of $E_2$. Then, for the second inequality
    \begin{align*}
        &\quad\mathbb{P}\big( \mathrm{CS}_c^i \subseteq 
            \mathcal{X}_m(c)
            \subseteq 
            \mathrm{CS}_c^o 
        \big)\\
        & = \mathbb{P}\Bigg(
            \inf_{\boldsymbol{x}_i \in \{d_n(\boldsymbol{x}_i)=0\}}\frac{\hat f_n(\boldsymbol{x}_i)-c}{\tau_n\sigma(\boldsymbol{x}_i)}\geq -a
            \bigcap 
            \forall \boldsymbol{x}_i \in \{d_n(\boldsymbol{x}_i)>0\}:\frac{\hat f_n(\boldsymbol{x}_i)-c}{\tau_n\sigma(\boldsymbol{x}_i)}\geq -a\\
            &\quad \quad \bigcap 
            \forall \boldsymbol{x}_i \in \{d_n(\boldsymbol{x}_i)<0\}:\frac{\hat f_n(\boldsymbol{x}_i)-c}{\tau_n\sigma(\boldsymbol{x}_i)}< a
        \Bigg) \quad \quad \dots\text{  Using lemma \ref{lemma:finite_equi}}\\
        & \geq \mathbb{P}\left(
            \inf_{\boldsymbol{x}_i \in \{d_n(\boldsymbol{x}_i)=0\}}\frac{\hat f_n(\boldsymbol{x}_i)-c}{\tau_n\sigma(\boldsymbol{x}_i)}\geq -a\right)
            +
            \mathbb{P}\left( 
            \forall \boldsymbol{x}_i \in \{d_n(\boldsymbol{x}_i)>0\}:\frac{\hat f_n(\boldsymbol{x}_i)-c}{\tau_n\sigma(\boldsymbol{x}_i)}\geq -a\right)-1\\
            &\quad +
            \mathbb{P}\left(
            \forall \boldsymbol{x}_i \in \{d_n(\boldsymbol{x}_i)<0\}:\frac{\hat f_n(\boldsymbol{x}_i)-c}{\tau_n\sigma(\boldsymbol{x}_i)}< a
        \right)-1 \quad \quad \dots \text{ Using Lemma \ref{lemma:intersection_bound}}\\
        & = \mathbb{P}\left(
            \inf_{\boldsymbol{x}_i \in \{d_n(\boldsymbol{x}_i)=0\}}G_n(\boldsymbol{x}_i)\geq -a\right)
            +
            \mathbb{P}\bigg(\forall \boldsymbol{x}_i \in \{d_n(\boldsymbol{x}_i)>0\}:G_n(\boldsymbol{x}_i)\geq -a-d_n(\boldsymbol{x}_i)\bigg)-1
            \\
            &\quad +\mathbb{P}\bigg(\forall \boldsymbol{x}_i \in \{d_n(\boldsymbol{x}_i)<0\}:G_n(\boldsymbol{x}_i)< a-d_n(\boldsymbol{x}_i)
        \bigg)-1\\
        & \geq \mathbb{P}\left(
            \inf_{\boldsymbol{x}_i \in \{d_n(\boldsymbol{x}_i)=0\}}G_n(\boldsymbol{x}_i)\geq -a\right)
            +
            \mathbb{P}\left(\forall \boldsymbol{x}_i \in \{d_n(\boldsymbol{x}_i)>0\}:G_n(\boldsymbol{x}_i)\geq -a-\inf_{\boldsymbol{x}_i \in \{d_n(\boldsymbol{x}_i)>0\}}d_n(\boldsymbol{x}_i)\right)-1
            \\
            &\quad +\mathbb{P}\left(\forall \boldsymbol{x}_i \in \{d_n(\boldsymbol{x}_i)<0\}:G_n(\boldsymbol{x}_i)< a+\inf_{\boldsymbol{x}_i \in \{d_n(\boldsymbol{x}_i)<0\}}|d_n(\boldsymbol{x}_i)|
        \right)-1\\
        & = \mathbb{P}\left(
            \inf_{\boldsymbol{x}_i \in \{d_n(\boldsymbol{x}_i)=0\}}G_n(\boldsymbol{x}_i)\geq -a\right)
            +
            \mathbb{P}\left(\inf_{\boldsymbol{x}_i \in \{d_n(\boldsymbol{x}_i)>0\}}G_n(\boldsymbol{x}_i)\geq -a-\inf_{\boldsymbol{x}_i \in \{d_n(\boldsymbol{x}_i)>0\}}d_n(\boldsymbol{x}_i)\right)-1
            \\
            &\quad +\mathbb{P}\left(\sup_{\boldsymbol{x}_i \in \{d_n(\boldsymbol{x}_i)<0\}}G_n(\boldsymbol{x}_i)< a+\inf_{\boldsymbol{x}_i \in \{d_n(\boldsymbol{x}_i)<0\}}|d_n(\boldsymbol{x}_i)|
        \right)-1
    \end{align*}
\end{proof}

Proof of Theorem \ref{theorem:asy_main_bound}:
Using Lemma \ref{lemma:main_bound} and take $n\rightarrow\infty$,
\begin{align*}
    & \lim_{n\rightarrow\infty}\mathbb{P}\left(
        \inf_{\boldsymbol{x}_i \in \{d_n(\boldsymbol{x}_i)=0\}}G_n(\boldsymbol{x}_i)\geq -a\right)\geq 
        \lim_{n\rightarrow\infty}\mathbb{P}\big( \mathrm{CS}_c^i \subseteq 
        \mathcal{X}_m(c)
        \subseteq 
        \mathrm{CS}_c^o
    \big)\geq \\
    & \lim_{n\rightarrow\infty}\mathbb{P}\left(
        \inf_{\boldsymbol{x}_i \in \{d_n(\boldsymbol{x}_i)=0\}}G_n(\boldsymbol{x}_i)\geq -a\right)
        +
        \lim_{n\rightarrow\infty}\mathbb{P}\left(\inf_{\boldsymbol{x}_i \in \{d_n(\boldsymbol{x}_i)>0\}}G_n(\boldsymbol{x}_i)\geq -a-\inf_{\boldsymbol{x}_i \in \{d_n(\boldsymbol{x}_i)>0\}}d_n(\boldsymbol{x}_i)\right)+
        \\
        &\quad \lim_{n\rightarrow\infty}\mathbb{P}\left(\sup_{\boldsymbol{x}_i \in \{d_n(\boldsymbol{x}_i)<0\}}G_n(\boldsymbol{x}_i)< a+\inf_{\boldsymbol{x}_i \in \{d_n(\boldsymbol{x}_i)<0\}}|d_n(\boldsymbol{x}_i)|
    \right)-2
\end{align*}
By Assumption \ref{assump:1}, $\forall \boldsymbol{x}_i \in \{d_n(\boldsymbol{x}_i)\neq 0\}: \lim_{n\rightarrow \infty}d_n(\boldsymbol{x}_i) \rightarrow \infty$ since $\tau_n\rightarrow 0$, and $G_n(\boldsymbol{x}_i)$ is almost surely bounded, then
\begin{align*}
    &\lim_{n\rightarrow\infty}\mathbb{P}\left(\inf_{\boldsymbol{x}_i \in \{d_n(\boldsymbol{x}_i)>0\}}G_n(\boldsymbol{x}_i)\geq -a-\inf_{\boldsymbol{x}_i \in \{d_n(\boldsymbol{x}_i)>0\}}d_n(\boldsymbol{x}_i)\right) = 1\\
    &\lim_{n\rightarrow\infty}\mathbb{P}\left(\sup_{\boldsymbol{x}_i \in \{d_n(\boldsymbol{x}_i)<0\}}G_n(\boldsymbol{x}_i)< a+\inf_{\boldsymbol{x}_i \in \{d_n(\boldsymbol{x}_i)<0\}}|d_n(\boldsymbol{x}_i)|\right)=1
\end{align*}
Therefore,
\begin{align*}
    \lim_{n\rightarrow\infty}\mathbb{P}\big( \mathrm{CS}_c^i \subseteq 
        \mathcal{X}_m(c)
        \subseteq 
        \mathrm{CS}_c^o
    \big) = 
    \lim_{n\rightarrow\infty}\mathbb{P}\left(
        \inf_{\boldsymbol{x}_i \in \{d_n(\boldsymbol{x}_i)=0\}}G_n(\boldsymbol{x}_i)\geq -a\right)
\end{align*}
\hfill\qedsymbol

Proof of Theorem \ref{theorem:finite_main_bound}:
    For the first inequality,
    \begin{align*}
        &\quad \quad\mathbb{P}\big( \mathrm{CS}_c^i \subseteq 
            \mathcal{X}_m(c)
            \subseteq 
            \mathrm{CS}_c^o 
        \big)\\
        & \leq \mathbb{P}\Bigg(
            \forall \boldsymbol{x}_i \in \{0 \leq d_n(\boldsymbol{x}_i)\leq e_1\}:\frac{\hat f_n(\boldsymbol{x}_i)-c}{\tau_n\sigma(\boldsymbol{x}_i)}\geq -a \\
            & \quad \quad \quad \bigcap 
            \forall \boldsymbol{x}_i \in \{-e_2 \leq d_n(\boldsymbol{x}_i) < 0 \}:\frac{\hat f_n(\boldsymbol{x}_i)-c}{\tau_n\sigma(\boldsymbol{x}_i)}< a
        \Bigg) \quad \quad \dots\text{  Using lemma \ref{lemma:finite_equi}}\\
        &\leq \mathbb{P}\Bigg(
            \forall \boldsymbol{x}_i \in \{0 \leq d_n(\boldsymbol{x}_i)\leq e_1\}:G_n(\boldsymbol{x}_i)\geq -a-d_n(\boldsymbol{x}_i) \\
            & \quad \quad \quad \bigcap 
            \forall \boldsymbol{x}_i \in \{-e_2 \leq d_n(\boldsymbol{x}_i) < 0 \}:G_n(\boldsymbol{x}_i)< a+|d_n(\boldsymbol{x}_i) |
        \Bigg)\\
        &\leq \mathbb{P}\Bigg(
            \forall \boldsymbol{x}_i \in \{0 \leq d_n(\boldsymbol{x}_i)\leq e_1\}:G_n(\boldsymbol{x}_i)\geq -a-\sup_{\boldsymbol{x}_i \in \{0\leq d_n(\boldsymbol{x}_i)\leq e_1\}}d_n(\boldsymbol{x}_i) \\
            & \quad \quad \quad \bigcap 
            \forall \boldsymbol{x}_i \in \{-e_2 \leq d_n(\boldsymbol{x}_i) < 0 \}:G_n(\boldsymbol{x}_i)< a+\sup_{\boldsymbol{x}_i \in \{-e_2 \leq d_n(\boldsymbol{x}_i) < 0 \}}|d_n(\boldsymbol{x}_i) |
        \Bigg)\\
        &\leq \mathbb{P}\Bigg(
            \inf_{\boldsymbol{x}_i \in \{0\leq d_n(\boldsymbol{x}_i)\leq e_1\}}G_n(\boldsymbol{x}_i)\geq -a-\sup_{\boldsymbol{x}_i \in \{0\leq d_n(\boldsymbol{x}_i)\leq e_1\}}d_n(\boldsymbol{x}_i) \\
            & \quad \quad \quad \bigcap 
            \sup_{\boldsymbol{x}_i \in \{-e_2 \leq d_n(\boldsymbol{x}_i) < 0 \}}G_n(\boldsymbol{x}_i)< a+\sup_{\boldsymbol{x}_i \in \{-e_2 \leq d_n(\boldsymbol{x}_i) < 0 \}}|d_n(\boldsymbol{x}_i) |
        \Bigg)
    \end{align*}
    Then, for the second inequality,
        \begin{align*}
        &\quad\mathbb{P}\big( \mathrm{CS}_c^i \subseteq 
            \mathcal{X}_m(c)
            \subseteq 
            \mathrm{CS}_c^o 
        \big)\\
        & = \mathbb{P}\Bigg(
            \forall \boldsymbol{x}_i \in \{0 \leq d_n(\boldsymbol{x}_i)\leq e_1\}:\frac{\hat f_n(\boldsymbol{x}_i)-c}{\tau_n\sigma(\boldsymbol{x}_i)}\geq -a \\
            & \quad \quad \quad \bigcap 
            \forall \boldsymbol{x}_i \in \{-e_2 \leq d_n(\boldsymbol{x}_i) < 0 \}:\frac{\hat f_n(\boldsymbol{x}_i)-c}{\tau_n\sigma(\boldsymbol{x}_i)}< a\\
            & \quad \quad \quad \bigcap 
            \forall \boldsymbol{x}_i \in \{e_1 < d_n(\boldsymbol{x}_i)\}:\frac{\hat f_n(\boldsymbol{x}_i)-c}{\tau_n\sigma(\boldsymbol{x}_i)}\geq -a\\
            & \quad \quad \quad \bigcap 
            \forall \boldsymbol{x}_i \in \{-e_2 > d_n(\boldsymbol{x}_i) \}:\frac{\hat f_n(\boldsymbol{x}_i)-c}{\tau_n\sigma(\boldsymbol{x}_i)}< a
        \Bigg)\quad \quad \dots\text{  Using lemma \ref{lemma:finite_equi}}\\
        & \geq \mathbb{P}\Bigg(
            \forall \boldsymbol{x}_i \in \{0 \leq d_n(\boldsymbol{x}_i)\leq e_1\}:\frac{\hat f_n(\boldsymbol{x}_i)-c}{\tau_n\sigma(\boldsymbol{x}_i)}\geq -a \\
            & \quad \quad \quad \bigcap 
            \forall \boldsymbol{x}_i \in \{-e_2 \leq d_n(\boldsymbol{x}_i) < 0 \}:\frac{\hat f_n(\boldsymbol{x}_i)-c}{\tau_n\sigma(\boldsymbol{x}_i)}< a
        \Bigg)
            +\\
            &\quad \sum_{\forall \boldsymbol{x}_i \in \{d_n(\boldsymbol{x}_i)> e_1\}}\mathbb{P}\Bigg(G_n(\boldsymbol{x}_i)\geq -a-d_n(\boldsymbol{x}_i)\Bigg) +\\ 
            & \quad \sum_{\forall \boldsymbol{x}_i \in \{d_n(\boldsymbol{x}_i) < -e_2\}} \mathbb{P}\Bigg(G_n(\boldsymbol{x}_i)< a+|d_n(\boldsymbol{x}_i)|
            \Bigg) - \\
            & \quad \mathrm{CARD}\Bigg\{\boldsymbol{x}_i \in \mathcal{X}_m: d_n(\boldsymbol{x}_i)> e_1\Bigg\}- \mathrm{CARD}\Bigg\{\boldsymbol{x}_i \in \mathcal{X}_m: d_n(\boldsymbol{x}_i) < -e_2\Bigg\} \quad \quad \dots \text{ Using Lemma \ref{lemma:intersection_bound}}\\
        & \geq \mathbb{P}\Bigg(
            \forall \boldsymbol{x}_i \in \{0 \leq d_n(\boldsymbol{x}_i)\leq e_1\}:G_n(\boldsymbol{x}_i)\geq -a-d_n(\boldsymbol{x}_i) \\
            & \quad \quad \quad \bigcap 
            \forall \boldsymbol{x}_i \in \{-e_2 \leq d_n(\boldsymbol{x}_i) < 0 \}:G_n(\boldsymbol{x}_i)< a+|d_n(\boldsymbol{x}_i) |
        \Bigg)
            +\\
            &\quad \sum_{\forall \boldsymbol{x}_i \in \{d_n(\boldsymbol{x}_i)> e_1\}}\mathbb{P}\Bigg(G_n(\boldsymbol{x}_i)\geq -a-d_n(\boldsymbol{x}_i)\Bigg) +\\ 
            & \quad \sum_{\forall \boldsymbol{x}_i \in \{d_n(\boldsymbol{x}_i) < -e_2\}} \mathbb{P}\Bigg(G_n(\boldsymbol{x}_i)< a+|d_n(\boldsymbol{x}_i)|
            \Bigg) - \\
            & \quad \mathrm{CARD}\Bigg\{\boldsymbol{x}_i \in \mathcal{X}_m: d_n(\boldsymbol{x}_i)> e_1\Bigg\}- \mathrm{CARD}\Bigg\{\boldsymbol{x}_i \in \mathcal{X}_m: d_n(\boldsymbol{x}_i) < -e_2\Bigg\}\\
        & \geq \mathbb{P}\Bigg(
            \inf_{\boldsymbol{x}_i \in \{0\leq d_n(\boldsymbol{x}_i)\leq e_1\}}G_n(\boldsymbol{x}_i)\geq -a-\inf_{\boldsymbol{x}_i \in \{0\leq d_n(\boldsymbol{x}_i)\leq e_1\}}d_n(\boldsymbol{x}_i) \\
            & \quad \quad \quad \bigcap 
            \sup_{\boldsymbol{x}_i \in \{-e_2 \leq d_n(\boldsymbol{x}_i) < 0 \}}G_n(\boldsymbol{x}_i)< a+\inf_{\boldsymbol{x}_i \in \{-e_2 \leq d_n(\boldsymbol{x}_i) < 0 \}}|d_n(\boldsymbol{x}_i) |
        \Bigg)
            +\\
            &\quad \sum_{\forall \boldsymbol{x}_i \in \{d_n(\boldsymbol{x}_i)> e_1\}}\mathbb{P}\Bigg(G_n(\boldsymbol{x}_i)\geq -a-d_n(\boldsymbol{x}_i)\Bigg) +\\ 
            & \quad \sum_{\forall \boldsymbol{x}_i \in \{d_n(\boldsymbol{x}_i) < -e_2\}} \mathbb{P}\Bigg(G_n(\boldsymbol{x}_i)< a+|d_n(\boldsymbol{x}_i)|
            \Bigg) - \\
            & \quad \mathrm{CARD}\Bigg\{\boldsymbol{x}_i \in \mathcal{X}_m: d_n(\boldsymbol{x}_i)> e_1\Bigg\}- \mathrm{CARD}\Bigg\{\boldsymbol{x}_i \in \mathcal{X}_m: d_n(\boldsymbol{x}_i) < -e_2\Bigg\}\\
        & \geq \mathbb{P}\Bigg(
            \sup_{\boldsymbol{x}_i \in \{-e_2 \leq d_n(\boldsymbol{x}_i) \leq e_1\}}|G_n(\boldsymbol{x}_i)|< a+\inf_{\boldsymbol{x}_i \in \{-e_2 \leq d_n(\boldsymbol{x}_i) \leq e_1 \}}|d_n(\boldsymbol{x}_i) |
        \Bigg)
            +\\
            &\quad \sum_{\forall \boldsymbol{x}_i \in \{d_n(\boldsymbol{x}_i)> e_1\}}\mathbb{P}\Bigg(G_n(\boldsymbol{x}_i)\geq -a-d_n(\boldsymbol{x}_i)\Bigg) +\\ 
            & \quad \sum_{\forall \boldsymbol{x}_i \in \{d_n(\boldsymbol{x}_i) < -e_2\}} \mathbb{P}\Bigg(G_n(\boldsymbol{x}_i)< a+|d_n(\boldsymbol{x}_i)|
            \Bigg) - \\
            & \quad \mathrm{CARD}\Bigg\{\boldsymbol{x}_i \in \mathcal{X}_m: d_n(\boldsymbol{x}_i)> e_1\Bigg\}- \mathrm{CARD}\Bigg\{\boldsymbol{x}_i \in \mathcal{X}_m: d_n(\boldsymbol{x}_i) < -e_2\Bigg\}
    \end{align*}

Proof of Corollary \ref{corollary:finite_sharp_bound}:
Using the conclusion from Theorem \ref{theorem:finite_main_bound}, we take $e_1$ and $e_2$ such that $\{0\leq d_n(\boldsymbol{x}_i)\leq e_1\}=\{\boldsymbol{x}_+\}$ and $\{-e_2\leq d_n(\boldsymbol{x}_i)< 0\}=\{\boldsymbol{x}_-\}$. By assumption \ref{assump:1}, there exists $n_1$ such that 
\begin{align*}
    \forall \boldsymbol{x}_i \in \mathcal{X}_m/\{\boldsymbol{x}_+,\boldsymbol{x}_-\}, \mathbb{P}(|G_n(\boldsymbol{x}_i)|\leq |d_n(\boldsymbol{x}_i)|)\geq 1-\frac{\delta}{m-2}
\end{align*}
Then, for all $a\geq 0$, we have
    \begin{align*}
        \delta &\geq \Bigg|\sum_{\forall \boldsymbol{x}_i \in \{c \leq f(\boldsymbol{x}_i)\}/x_+}\mathbb{P}\Bigg(G_n(\boldsymbol{x}_i)\geq -a-d_n(\boldsymbol{x}_i)\Bigg) +\\ 
        & \quad \sum_{\forall \boldsymbol{x}_i \in \{c > f(\boldsymbol{x}_i)\}/x_{-}} \mathbb{P}\Bigg(G_n(\boldsymbol{x}_i)< a+|d_n(\boldsymbol{x}_i)|
        \Bigg) -  (m-2)\Bigg|
    \end{align*}
Thus,
    \begin{align*}
        & \mathbb{P}\Bigg(
        G_n(\boldsymbol{x}_+)\geq -a-d_n(\boldsymbol{x}_+)
        \bigcap 
        G_n(\boldsymbol{x}_-)< a+|d_n(\boldsymbol{x}_-)|
        \Bigg) \geq \\
        &\mathbb{P}\Bigg( 
            \mathrm{CS}_c^i \subseteq 
            \mathcal{X}_m(c)
            \subseteq 
            \mathrm{CS}_c^o
        \Bigg)\geq \\
        & \mathbb{P}\Bigg(
        G_n(\boldsymbol{x}_+)\geq -a-d_n(\boldsymbol{x}_+)
        \bigcap 
        G_n(\boldsymbol{x}_-)< a+|d_n(\boldsymbol{x}_-)|
        \Bigg) - \delta
    \end{align*}

By assumption \ref{assump:1}, we have 
\begin{align*}
    &\forall \boldsymbol{x}_i \in \mathcal{X}_m, 
    \lim_{n\rightarrow \infty}d_n(\boldsymbol{x}_i)\rightarrow \infty,\\ &\lim_{n\rightarrow \infty}\mathbb{P}(|G_n(\boldsymbol{x}_i)|\leq |d_n(\boldsymbol{x}_i)|)=1,
\end{align*}
and using Theorem \ref{theorem:finite_main_bound}, we have for all $a\geq 0$,
    \begin{align*}
        \lim_{n\rightarrow \infty}\mathbb{P}\big( 
        \mathrm{CS}_c^i \subseteq 
        \mathcal{X}_m(c)
        \subseteq 
        \mathrm{CS}_c^o\big)=
        \lim_{n\rightarrow \infty}\mathbb{P}\big(
        G_n(\boldsymbol{x}_+)\geq -a-d_n(\boldsymbol{x}_+)
        \cap 
        G_n(\boldsymbol{x}_-)< a+|d_n(\boldsymbol{x}_-)|
        \big)=1.
    \end{align*}

\end{document}